\documentclass[journal]{IEEEtran}

\ifCLASSINFOpdf

\else

\fi

\usepackage[numbers,sort,compress]{natbib}
\usepackage{graphicx}
\usepackage{amsmath}
\usepackage{amssymb}
\usepackage{booktabs}
\usepackage{multirow}
\usepackage{times}
\usepackage{setspace}
\usepackage{subfigure}
\usepackage{graphicx}
\usepackage{amsfonts}
\usepackage{comment}
\usepackage{color}
\usepackage{xcolor}
\usepackage{tabularx}
\usepackage{makecell}
\usepackage{enumerate}
\usepackage[T1]{fontenc}
\usepackage{hyperref}
\usepackage{bm}
\usepackage{changepage}
\usepackage{lineno}
\usepackage{algorithmic}
\usepackage{array}
\usepackage{dsfont}
\usepackage[ruled,lined,linesnumbered]{algorithm2e}
\usepackage{multirow}
\usepackage{multicol}
\usepackage{hhline}

\graphicspath{ {figures/} }

\newcommand{\HH}{\mathcal{H}}

\ifCLASSINFOpdf
\else
\fi

\newcommand*\rot{\rotatebox{90}}

\begin{document}
%
\title{A Review of Single-Source Deep Unsupervised Visual Domain Adaptation}

\author{Sicheng~Zhao,~\IEEEmembership{Senior Member, IEEE},~Xiangyu~Yue,~Shanghang~Zhang,~Bo~Li,~Han~Zhao,~Bichen~Wu,~Ravi~Krishna,~Joseph~E.~Gonzalez,~Alberto~L.~Sangiovanni-Vincentelli,~\IEEEmembership{Fellow, IEEE}, Sanjit A. Seshia~\IEEEmembership{Fellow, IEEE},~and~Kurt~Keutzer,~\IEEEmembership{Fellow, IEEE}
\thanks{Copyright (c) 2013 IEEE. Personal use of this material is permitted. However, permission to use this material for any other purposes must be obtained from the IEEE by sending a request to pubs-permissions@ieee.org.}
\thanks{Manuscript received July 17, 2019, revised May 15, 2020 and September 18, 2020.  This work was supported by Berkeley DeepDrive. Corresponding authors: Xiangyu Yue, Shanghang Zhang.}
\thanks{S. Zhao, X. Yue, S. Zhang, B. Li, B. Wu, R. Krishna, J. E. Gonzalez, A. L. Sangiovanni-Vincentelli, S. A. Seshia, and K. Keutzer are with the Department of Electrical Engineering and Computer Sciences, University of California, Berkeley, CA 94720, USA (e-mail: schzhao@gmail.com, xyyue@eecs.berkeley.edu, shz@eecs.berkeley.edu, drluodian@gmail.com, bichen@berkeley.edu, ravi.krishna@berkeley.edu, jegonzal@berkeley.edu, alberto@berkeley.edu, sseshia@berkeley.edu,  keutzer@berkeley.edu).}
\thanks{H. Zhao is with School of Computer Science, Carnegie Mellon University, Pittsburgh, PA 15213, USA (e-mail: han.zhao@cs.cmu.edu).}
}

\markboth{Under review}
{Shell Zhao{\textit{et al.}}: Single-Source Unsupervised Visual Domain Adaptation}

\maketitle

\begin{abstract}
Large-scale labeled training datasets have enabled deep neural networks to excel across a wide range of benchmark vision tasks.
However, in many applications, it is prohibitively expensive and time-consuming to obtain large quantities of labeled data. To cope with limited labeled training data, many have attempted to directly apply models trained on a large-scale labeled source domain to another sparsely labeled or unlabeled target domain. Unfortunately, direct transfer across domains often performs poorly due to the presence of \emph{domain shift} or \emph{dataset bias}. Domain adaptation is a machine learning paradigm that aims to learn a model from a source domain that can perform well on a different (but related) target domain. In this paper, we review the latest single-source deep unsupervised domain adaptation methods focused on visual tasks and discuss new perspectives for future research. We begin with the definitions of different domain adaptation strategies and the descriptions of existing benchmark datasets. We then summarize and compare different categories of single-source unsupervised domain adaptation methods, including discrepancy-based methods, adversarial discriminative methods, adversarial generative methods, and self-supervision-based methods. Finally, we discuss future research directions with challenges and possible solutions.
\end{abstract}


\IEEEpeerreviewmaketitle

\section{Introduction}
\label{sec:Introduction}

In the last decade, deep neural networks (DNNs) have achieved significant progress in various computer vision tasks where large-scale labeled training data are available. For example, the classification error of the ``Classification + localization with provided training data'' task in the Large Scale Visual Recognition Challenge was reduced from 0.28 in 2010 to 0.022 in 2017\footnote{\url{http://image-net.org/challenges/LSVRC/2017}}, even outperforming humans.
However, in many applications, it is difficult to obtain a large amount of labels, as manual annotation is expensive and time-consuming.

\begin{figure}[!t]
\begin{center}
\centering \includegraphics[width=0.95\linewidth]{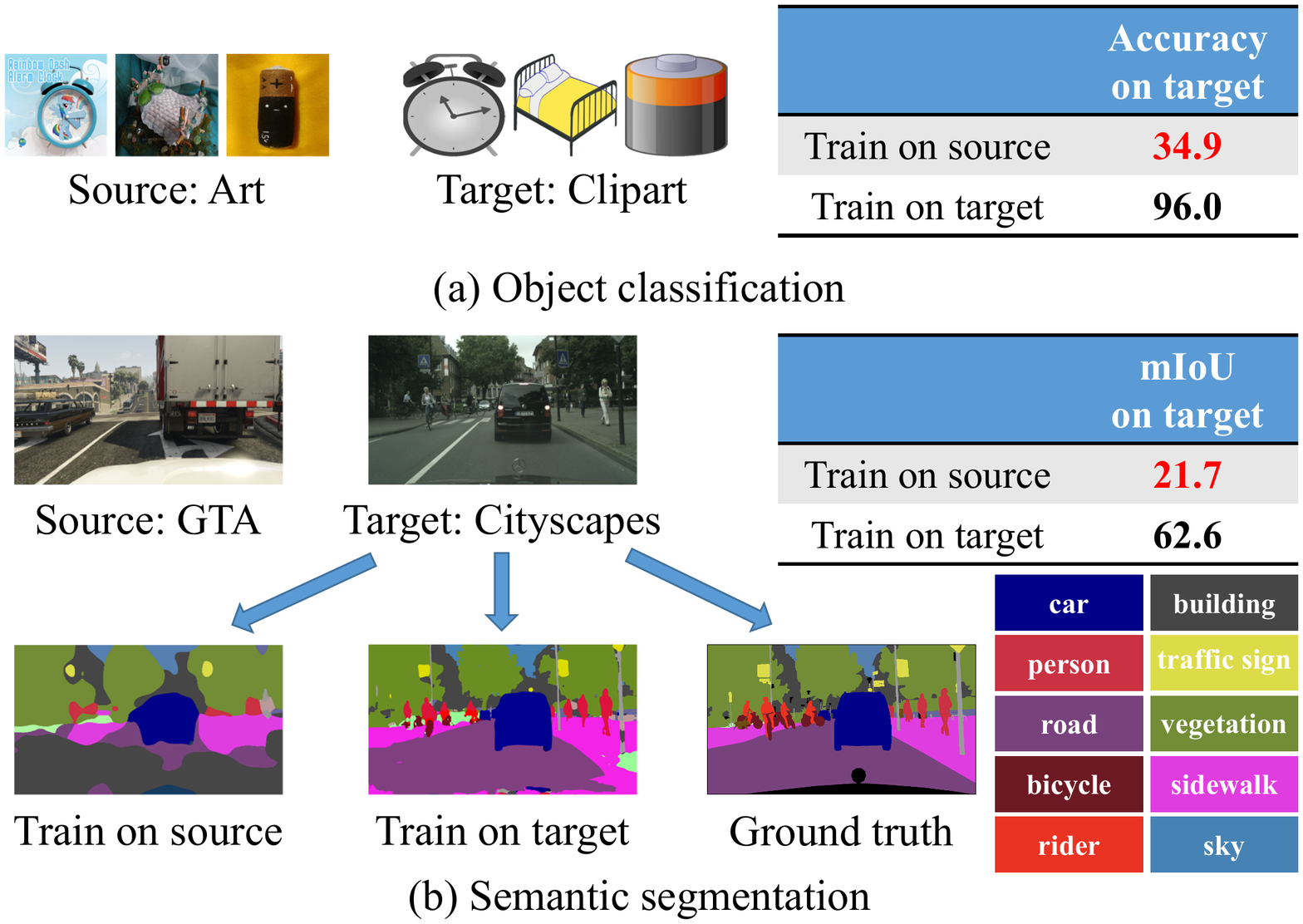}
\caption{An example of \emph{domain shift}. For both image-level object classification and pixel-wise semantic segmentation tasks, direct transfer of the models trained on the labeled source domain to the unlabeled target domain results in a dramatic performance drop.}
\label{fig:DatasetBias}
\end{center}
\end{figure}

An alternative solution is to train a model on another related large-scale source domain with labels (\textit{e.g.} a simulation domain) and apply it to the unlabeled target domain (\textit{e.g.} a real-world domain). However, due to the presence of \emph{domain shift} or \emph{dataset bias}~\cite{torralba2011unbiased}, such a direct transfer might not perform well, as shown in Figure~\ref{fig:DatasetBias}.

One may argue that the pre-trained source models can be fine-tuned in the target domain.
However, fine-tuning still requires considerable quantities of labeled training data, which may be not available for many applications.
For example, in fine-grained recognition, only experts are able to provide reliable labeled data~\cite{gebru2017fine}; in semantic segmentation, it takes about 90 minutes to label each image in the Cityscapes dataset~\cite{cordts2016cityscapes}; in autonomous driving, the substantial traffic data obtained with different sensors, such as 3D LiDAR point clouds, are difficult to label~\cite{wu2017squeezeseg, yue2018lidar}; in affective image content analysis, the perceived emotions are subjective and personalized across viewers~\cite{zhao2018emotiongan,zhao2019cycleemotiongan}.

\begin{figure*}[!t]
\begin{center}
\subfigure[Discrepancy-based methods]{
\includegraphics[width=0.46\linewidth]{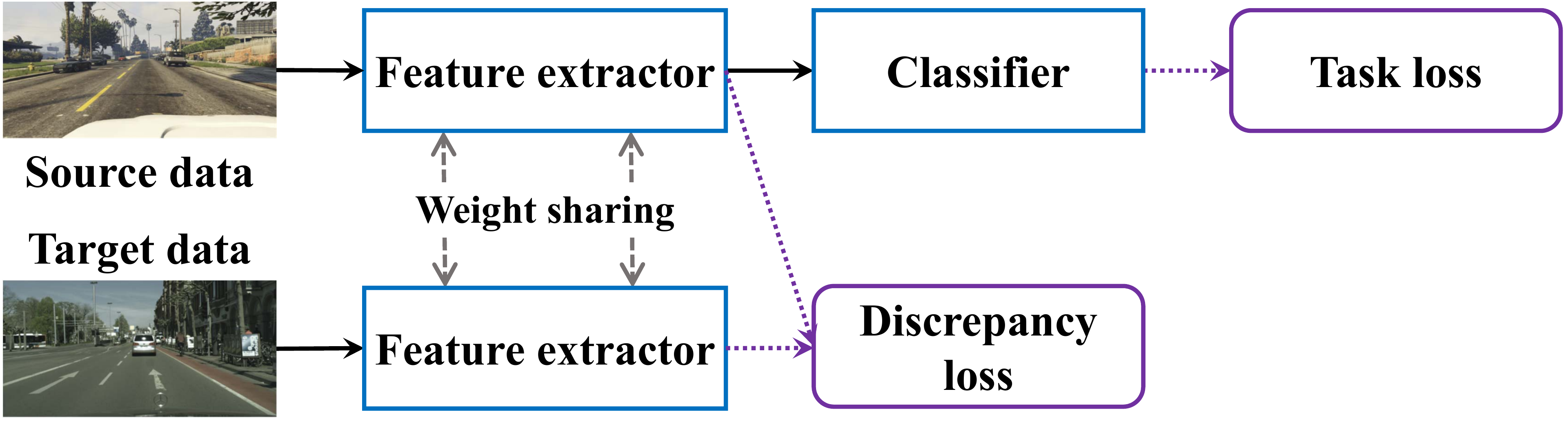}
}
\subfigure[Adversarial discriminative methods]{
\includegraphics[width=0.46\linewidth]{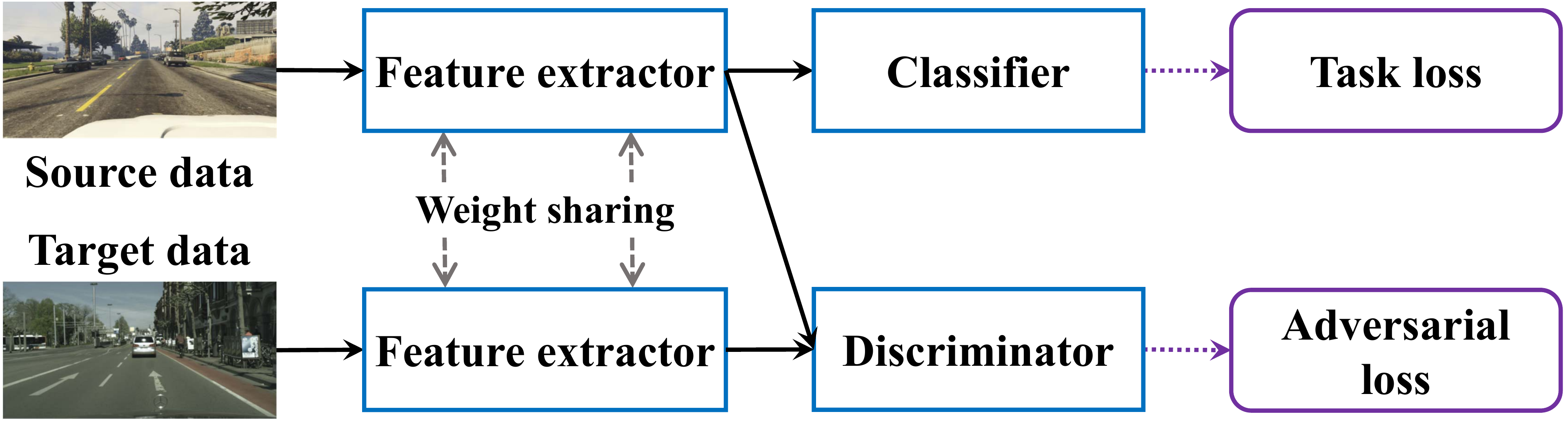}
}
\subfigure[Adversarial generative methods]{
\includegraphics[width=0.46\linewidth]{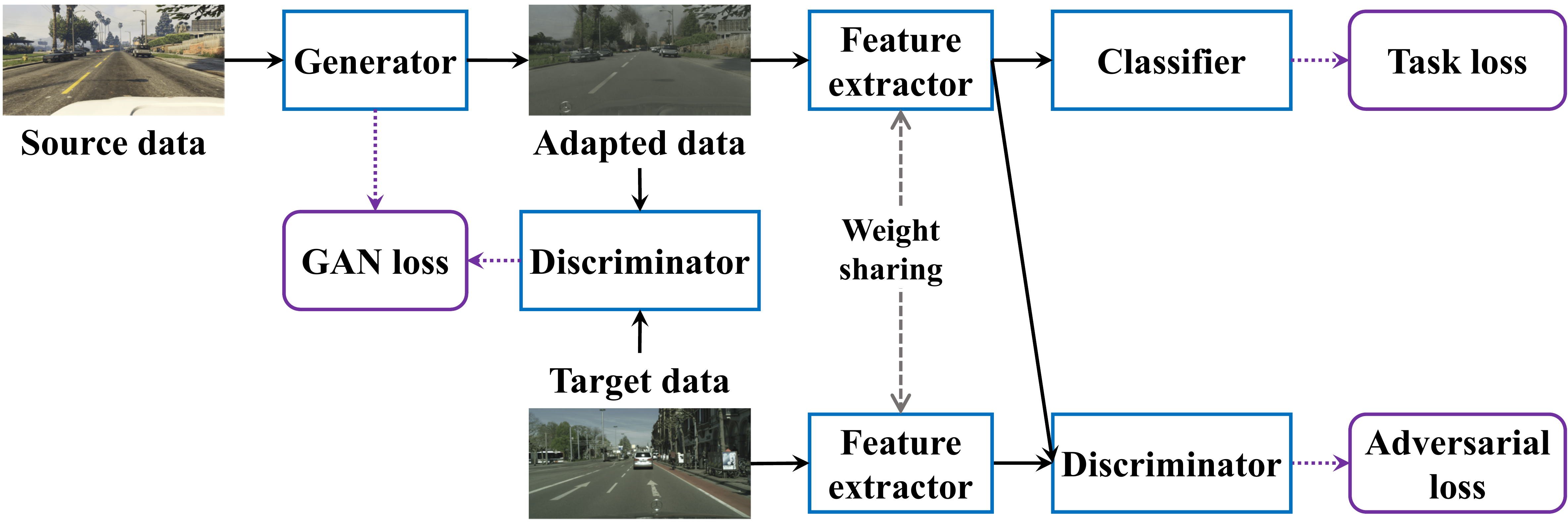}
}
\subfigure[Self-supervision-based methods]{
\includegraphics[width=0.46\linewidth]{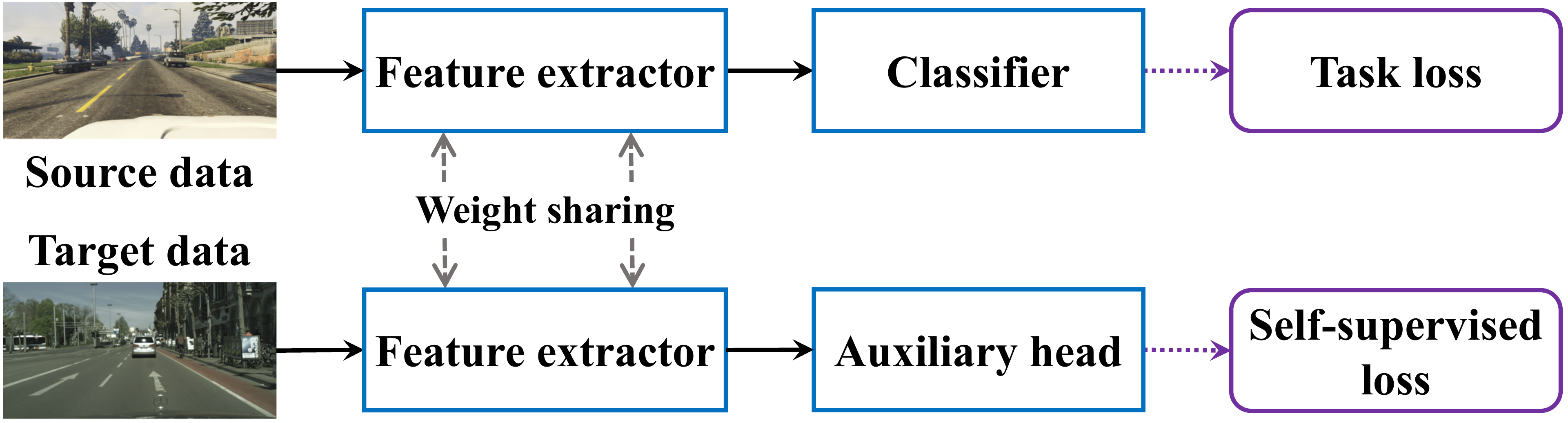}
}
\caption{Classification of widely employed framework of different single-source deep unsupervised domain adaptation (DUDA) pipelines. Most existing methods can be obtained by employing different component values, slightly changing the architecture, or combining different pipelines.}
\label{fig:Framework}
\end{center}
\end{figure*}

\paragraph{Domain Adaptation in context of other sample-efficient learning methods}
Domain adaptation techniques were introduced to addresses the domain shift between source and target domains~\cite{you2019towards} and for this reason,
they have recently attracted significant interest in both academia and industry. 
\textit{Domain adaptation (DA)}, also known as \textit{domain transfer}, is a specialized form of transfer learning~\cite{pan2010survey} that aims to learn a model from a labeled source domain that can  generalize well to a different (but related) unlabeled or sparsely labeled target domain. It belongs to the sample-efficient learning class~\cite{luo2017label}, together with  zero-shot learning, few-shot learning, and self-supervised learning.  We briefly compare domain adaptation with other methods in this category. While the unsupervised domain adaptation (UDA) does not require the annotations of target data, it usually needs a sufficient number of unlabeled target samples to train the model~\cite{tzeng2017adversarial}. Compared to UDA, zero-shot learning does not need either the target data annotations or the unlabeled target samples~\cite{romera2015embarrassingly,ni2019dual}. However, existing methods often require some auxiliary information, such as the attributes of the images, or the description of the classes~\cite{socher2013zero,song2019generalized}. Further, zero-shot learning is trained on known/seen classes and tested on unknown/unseen classes, which demands the model to generalize from known/seen classes to unknown/unseen classes. Since the known/seen classes and the unknown/unseen classes are from different distributions, there is no concept of domain shift in zero-shot learning. Different from zero-shot learning, DA deals with the same learning tasks on different domains. Taking image classification as an example, both source data and target data have the same classes. Few-shot learning shares similar setting with zero-shot learning. The difference is that few-shot learning has a few (\textit{e.g.} 5 or 10) annotated samples for the unknown/unseen classes~\cite{snell2017prototypical,ravi2017optimization,sung2018learning}. Few-shot learning and zero-shot learning can also be grouped as low-shot learning. 

Self-supervised learning (SSL) is a learning paradigm that captures the intrinsic patterns and properties of input data without using human-provided labels~\cite{chen2020simple}. The basic idea of SSL is to construct some auxiliary tasks solely based on the data itself without using human-annotated labels and force the network to learn meaningful representations by performing the auxiliary tasks well. Typical self-supervised learning approaches generally involve two aspects: constructing auxiliary tasks and defining loss functions~\cite{he2019momentum}. The auxiliary tasks are designed to encourage the model to learn meaningful representations of input data. The loss functions are defined to measure the difference between a model’s prediction and a fixed target, the similarities of sample pairs in a representation space (\textit{e.g.}, contrastive loss), or the difference between probability distributions (\textit{e.g.}, adversarial loss). Compared with domain adaptation, SSL does not specifically address the domain shift problem between different domains.

\paragraph{Domain Adaptation Challenges}
\label{ssec:challenge}
Albeit DA is a very effective method, it is not without blemish. The main challenge for single-source UDA is domain shift~\cite{torralba2011unbiased}, \textit{i.e.}, the difference between the source and target distributions that leads to unreliable predictions on the target domain. Typically, three types of domain shift are considered: covariate shift, label shift, and concept drift (see Section~\ref{sec:ProblemDefinition} for details).

As Figure~\ref{fig:DatasetBias} shows, the presence of domain shift causes the direct transfer of models trained on the source domain to perform poorly on the target domain. Figure~\ref{fig:DatasetBias}~(a) shows an example of image-level object classification on the Office-Home dataset~\cite{venkateswara2017deep}. When training a ResNet-50 model~\cite{he2016deep} on the target Clipart domain, we can obtain a promising 96.0\% object classification accuracy. However, if we train the model on the source Art domain and directly test it on the target domain, the accuracy significantly drops to 34.9\%.

Figure~\ref{fig:DatasetBias}~(b) shows an example of simulation-to-real adaptation, which is a more realistic application with unlimited synthetic labeled data created by graphics and simulation infrastructure. For example, CARLA\footnote{\url{http://www.carla.org}}, GTA-V\footnote{\url{https://www.rockstargames.com/V}}, and Autoware.AI\footnote{\url{https://www.autoware.ai/}} are three popular simulators for autonomous driving research. While there are ongoing efforts to make simulations more realistic, it is very difficult to model all the characteristics of real data~\cite{shrivastava2017learning}. Using the FCN model~\cite{long2015fully} to conduct pixel-wise segmentation on a real target dataset Cityscapes~\cite{cordts2016cityscapes}, direct transfer from synthetic GTA~\cite{richter2016playing} only obtains a mean intersection-over-union (mIoU) of 21.7\%, which is much lower than the mIoU 62.6\% of the model trained on real Cityscapes.

\paragraph{Focus of this Survey and Comparison with Other Surveys}
\label{ssec:scope}
There are many different domain adaptation settings (see Section~\ref{sec:ProblemDefinition} for details). Our survey focuses on the most prevalent one: single-source, single-target, homogeneous, and closed set adaptation without target labels. In this setting, there is one fully labeled source domain and one unlabeled target domain within the same modality, and the source and target domains share the same label set.

The early unsupervised domain adaptation (UDA) methods were mainly non-deep approaches, which aimed to match the feature distributions between the source domain and the target domain. Single-source UDA methods can be divided into two categories~\cite{patel2015visual,csurka2017domain}: sample re-weighting~\cite{huang2007correcting,gong2013connecting} and intermediate subspace transformation~\cite{gopalan2011domain,gong2012geodesic,ni2013subspace,fernando2013unsupervised,gopalan2014unsupervised}.

With the advent of deep learning, the emphasis has shifted to end-to-end learning domain-invariant features.
Typically, for single-source deep UDA (DUDA)~\cite{das2018graph,you2019towards}, a conjoined architecture with two streams is employed to represent the models for the source and target domains, respectively~\cite{zhuo2017deep}. Besides the traditional task loss, such as cross-entropy loss for classification, based on the labeled source data, DUDA models are usually trained jointly with another loss to deal with domain shift, such as a discrepancy loss, adversarial loss, or self-supervision loss. The single-source DUDA methods are divided into four categories based on domain shift loss and generative/discriminative settings, as shown in Figure~\ref{fig:Framework}.

There have been several surveys on domain adaptation and transfer learning. In particular: \cite{pan2010survey} covers different transfer learning paradigms, such as self-taught learning and multi-task learning, but the domain adaptation part is not comprehensive; \cite{patel2015visual} deals for the most part with early methods with little discussion devoted to recent deep learning-based methods; \cite{csurka2017domain} covers almost all categories of domain adaptation methods briefly but comparisons are scant; \cite{kouw2019review,zhang2019transfer}  focus on shallow methods and only a few deep methods are reviewed; \cite{sun2015survey,zhao2020multi} analyze  multi-source domain adaptation respectively focusing on  shallow  and deep methods. Similar to \cite{wang2018deep,tan2018survey,wilson2018survey}, we focus on deep domain adaptation, but use a different taxonomy that provides different insights. There are also some blogs summarizing recent papers on different transfer learning methods\footnote{\url{https://github.com/jindongwang/transferlearning}} and domain adaptation\footnote{\url{https://github.com/zhaoxin94/awesome-domain-adaptation}} strategies. Compared to existing surveys and blogs, our survey has the following advantages/contributions: (1) we cover and compare the latest methods on deep unsupervised domain adaptation; (2) we provide an analysis that includes advantages/disadvantages of different categories and differences/connections among different methods in each category with summarizing tables; (3) we give suggestions and prospects for future directions to explore; (4) we systematically compare the results of existing methods on popular benchmark datasets; and (5) we discuss some important aspects that are overlooked in previous surveys, such as label shift and theory.



\begin{table*}[!t]
\centering
\caption{Classification of domain adaptation (DA) strategies.}
\resizebox{0.95\textwidth}{!}{%
\begin{tabular}
{ccccc}
\toprule
\textbf{Standard} & \textbf{Classification} & \textbf{Definition} & \textbf{Short description} &  \textbf{Representative methods}\\
\hline
\multirow{3}{*}{target label}   & unsupervised DA &  $N_{TL}=0$ & target data is fully unlabeled  & 
\cite{zhuo2017deep,shrivastava2017learning,tzeng2017adversarial,wu2019squeezesegv2,chen2018domain,long2015learning,sun2017correlation,rozantsev2018beyond,zellinger2017central,chen2020homm,kang2019contrastive,long2017deep,li2018adaptive,cariucci2017autodial,lee2019sliced,roy2019unsupervised,hoffman2016fcns,ganin2016domain,chen2017no,hong2018conditional,tsai2018learning,long2018conditional,cicek2019unsupervised,hu2020panda,liu2019transferable,xu2019adversarial,du2020dual,chen2020adversarial,pan2019transferrable,bousmalis2017unsupervised,hoffman2018cycada,kang2018deep,sankaranarayanan2018generate,tran2019gotta,li2019cycle,wu2018dcan,liu2017unsupervised,russo2018source,ghifary2016deep,bousmalis2016domain,sun2019unsupervised,carlucci2019domain,xu2019self,feng2019self,kim2020cross,achituve2020self,tzeng2018splat,zhang2018fully,murez2018image,zhao2019learning,combes2020domain,lipton2018detecting,tan2019generalized,azizzadenesheli2019regularized,zou2018unsupervised,zou2019confidence,laine2017temporal,tarvainen2017mean,french2018self,athiwaratkun2019there,zhang2017curriculum,wang2018visual,saito2018maximum,chen2018road,wang2019weakly,shu2019transferable,wang2011heterogeneous,li2014learning,hubert2016learning,li2018heterogeneous,li2019heterogeneous,yao2019heterogeneous,yu2018multi,gholami2020unsupervised,panareda2017open,saito2018open,liu2019separate,cao2018partial,zhang2018importance,cao2018partialadversarial,cao2019learning,you2019universal,cai2019exploring,khodabandeh2019robust,zhu2019adapting,he2019multi,saito2019strong,kim2019diversify,xie2019multi,hsu2020progressive,zheng2020cross} \\
 & fully supervised DA & $N_{TL}=N_T$ & target data is fully labeled & \cite{wang2018deep} \\
 & semi-supervised DA & $0<N_{TL}<N_T$ & target data is partially unlabeled  &  \cite{li2014learning,tzeng2015simultaneous,ballester2019semi,saito2019semi} \\
\hline
\multirow{2}{*}{source label}   & strongly-supervised DA &  $N_{SL}=N_S$ & source data is fully/strongly labeled &  
\cite{zhuo2017deep,shrivastava2017learning,tzeng2017adversarial,wu2019squeezesegv2,chen2018domain,long2015learning,sun2017correlation,rozantsev2018beyond,zellinger2017central,chen2020homm,kang2019contrastive,long2017deep,li2018adaptive,cariucci2017autodial,lee2019sliced,roy2019unsupervised,hoffman2016fcns,ganin2016domain,chen2017no,hong2018conditional,tsai2018learning,long2018conditional,cicek2019unsupervised,hu2020panda,liu2019transferable,xu2019adversarial,du2020dual,chen2020adversarial,pan2019transferrable,bousmalis2017unsupervised,hoffman2018cycada,kang2018deep,sankaranarayanan2018generate,tran2019gotta,li2019cycle,wu2018dcan,liu2017unsupervised,russo2018source,ghifary2016deep,bousmalis2016domain,sun2019unsupervised,carlucci2019domain,xu2019self,feng2019self,kim2020cross,achituve2020self,tzeng2018splat,zhang2018fully,murez2018image,zhao2019learning,combes2020domain,lipton2018detecting,tan2019generalized,azizzadenesheli2019regularized,zou2018unsupervised,zou2019confidence,laine2017temporal,tarvainen2017mean,french2018self,athiwaratkun2019there,zhang2017curriculum,wang2018visual,saito2018maximum,chen2018road,li2014learning,tzeng2015simultaneous,ballester2019semi,saito2019semi,wang2011heterogeneous,hubert2016learning,li2018heterogeneous,li2019heterogeneous,yao2019heterogeneous,sun2015survey,zhao2020multi,duan2009domain,sun2011two,duan2012exploiting,chattopadhyay2012multisource,duan2012domain,yang2007cross,schweikert2009empirical,xu2012multi,sun2013bayesian,bhatt2016multi,xu2018deep,zhao2018adversarial,hoffman2018algorithms,peng2019moment,zhao2019multi,zhao2020distilling,lin2020multi,yu2018multi,gholami2020unsupervised,panareda2017open,saito2018open,liu2019separate,cao2018partial,zhang2018importance,cao2018partialadversarial,cao2019learning,you2019universal,cai2019exploring,khodabandeh2019robust,zhu2019adapting,he2019multi,saito2019strong,kim2019diversify,xie2019multi,hsu2020progressive,zheng2020cross}\\
 & weakly-supervised DA & $N_{SL}<N_T$ & source data is weakly labeled  & \cite{wang2019weakly,shu2019transferable} \\
\hline
\multirow{2}{*}{homogeneity}   & homogeneous DA &  $d_S=d_T$ & source and target data are observed in the same space &  
\cite{zhuo2017deep,shrivastava2017learning,tzeng2017adversarial,wu2019squeezesegv2,chen2018domain,long2015learning,sun2017correlation,rozantsev2018beyond,zellinger2017central,chen2020homm,kang2019contrastive,long2017deep,li2018adaptive,cariucci2017autodial,lee2019sliced,roy2019unsupervised,hoffman2016fcns,ganin2016domain,chen2017no,hong2018conditional,tsai2018learning,long2018conditional,cicek2019unsupervised,hu2020panda,liu2019transferable,xu2019adversarial,du2020dual,chen2020adversarial,pan2019transferrable,bousmalis2017unsupervised,hoffman2018cycada,kang2018deep,sankaranarayanan2018generate,tran2019gotta,li2019cycle,wu2018dcan,liu2017unsupervised,russo2018source,ghifary2016deep,bousmalis2016domain,sun2019unsupervised,carlucci2019domain,xu2019self,feng2019self,kim2020cross,achituve2020self,tzeng2018splat,zhang2018fully,murez2018image,zhao2019learning,combes2020domain,lipton2018detecting,tan2019generalized,azizzadenesheli2019regularized,zou2018unsupervised,zou2019confidence,laine2017temporal,tarvainen2017mean,french2018self,athiwaratkun2019there,zhang2017curriculum,wang2018visual,saito2018maximum,chen2018road,tzeng2015simultaneous,ballester2019semi,saito2019semi,wang2019weakly,shu2019transferable,duan2009domain,sun2011two,duan2012exploiting,chattopadhyay2012multisource,duan2012domain,yang2007cross,schweikert2009empirical,xu2012multi,sun2013bayesian,bhatt2016multi,xu2018deep,zhao2018adversarial,hoffman2018algorithms,peng2019moment,zhao2019multi,zhao2020distilling,lin2020multi,yu2018multi,gholami2020unsupervised,panareda2017open,saito2018open,liu2019separate,cao2018partial,zhang2018importance,cao2018partialadversarial,cao2019learning,you2019universal,cai2019exploring,khodabandeh2019robust,zhu2019adapting,he2019multi,saito2019strong,kim2019diversify,xie2019multi,hsu2020progressive,zheng2020cross}\\
 & heterogeneous DA & $d_S\neq d_T$ & source and target data are observed in different spaces  & \cite{wang2011heterogeneous,li2014learning,hubert2016learning,li2018heterogeneous,li2019heterogeneous,yao2019heterogeneous} \\
\hline
\multirow{2}{*}{\#sources}   & single-source DA &  $N_S=1$ & there is only one source domain &  
\cite{zhuo2017deep,shrivastava2017learning,tzeng2017adversarial,wu2019squeezesegv2,chen2018domain,long2015learning,sun2017correlation,rozantsev2018beyond,zellinger2017central,chen2020homm,kang2019contrastive,long2017deep,li2018adaptive,cariucci2017autodial,lee2019sliced,roy2019unsupervised,hoffman2016fcns,ganin2016domain,chen2017no,hong2018conditional,tsai2018learning,long2018conditional,cicek2019unsupervised,hu2020panda,liu2019transferable,xu2019adversarial,du2020dual,chen2020adversarial,pan2019transferrable,bousmalis2017unsupervised,hoffman2018cycada,kang2018deep,sankaranarayanan2018generate,tran2019gotta,li2019cycle,wu2018dcan,liu2017unsupervised,russo2018source,ghifary2016deep,bousmalis2016domain,sun2019unsupervised,carlucci2019domain,xu2019self,feng2019self,kim2020cross,achituve2020self,tzeng2018splat,zhang2018fully,murez2018image,zhao2019learning,combes2020domain,lipton2018detecting,tan2019generalized,azizzadenesheli2019regularized,zou2018unsupervised,zou2019confidence,laine2017temporal,tarvainen2017mean,french2018self,athiwaratkun2019there,zhang2017curriculum,wang2018visual,saito2018maximum,chen2018road,li2014learning,tzeng2015simultaneous,ballester2019semi,saito2019semi,wang2019weakly,shu2019transferable,yu2018multi,gholami2020unsupervised,panareda2017open,saito2018open,liu2019separate,cao2018partial,zhang2018importance,cao2018partialadversarial,cao2019learning,you2019universal,wang2011heterogeneous,hubert2016learning,li2018heterogeneous,li2019heterogeneous,yao2019heterogeneous,cai2019exploring,khodabandeh2019robust,zhu2019adapting,he2019multi,saito2019strong,kim2019diversify,xie2019multi,hsu2020progressive,zheng2020cross}\\
 & multi-source DA & $N_S>1$ & there are multiple source domains  & \cite{sun2015survey,zhao2020multi,duan2009domain,sun2011two,duan2012exploiting,chattopadhyay2012multisource,duan2012domain,yang2007cross,schweikert2009empirical,xu2012multi,sun2013bayesian,bhatt2016multi,xu2018deep,zhao2018adversarial,hoffman2018algorithms,peng2019moment,zhao2019multi,zhao2020distilling,lin2020multi} \\
\hline
\multirow{2}{*}{\#targets}   & single-target DA &  $N_T=1$ & there is only one target domain &  
\cite{zhuo2017deep,shrivastava2017learning,tzeng2017adversarial,wu2019squeezesegv2,chen2018domain,long2015learning,sun2017correlation,rozantsev2018beyond,zellinger2017central,chen2020homm,kang2019contrastive,long2017deep,li2018adaptive,cariucci2017autodial,lee2019sliced,roy2019unsupervised,hoffman2016fcns,ganin2016domain,chen2017no,hong2018conditional,tsai2018learning,long2018conditional,cicek2019unsupervised,hu2020panda,liu2019transferable,xu2019adversarial,du2020dual,chen2020adversarial,pan2019transferrable,bousmalis2017unsupervised,hoffman2018cycada,kang2018deep,sankaranarayanan2018generate,tran2019gotta,li2019cycle,wu2018dcan,liu2017unsupervised,russo2018source,ghifary2016deep,bousmalis2016domain,sun2019unsupervised,carlucci2019domain,xu2019self,feng2019self,kim2020cross,achituve2020self,tzeng2018splat,zhang2018fully,murez2018image,zhao2019learning,combes2020domain,lipton2018detecting,tan2019generalized,azizzadenesheli2019regularized,zou2018unsupervised,zou2019confidence,laine2017temporal,tarvainen2017mean,french2018self,athiwaratkun2019there,zhang2017curriculum,wang2018visual,saito2018maximum,chen2018road,li2014learning,tzeng2015simultaneous,ballester2019semi,saito2019semi,wang2019weakly,shu2019transferable,duan2009domain,sun2011two,duan2012exploiting,chattopadhyay2012multisource,duan2012domain,yang2007cross,schweikert2009empirical,xu2012multi,sun2013bayesian,bhatt2016multi,xu2018deep,zhao2018adversarial,hoffman2018algorithms,peng2019moment,zhao2019multi,zhao2020distilling,lin2020multi,panareda2017open,saito2018open,liu2019separate,cao2018partial,zhang2018importance,cao2018partialadversarial,cao2019learning,you2019universal,wang2011heterogeneous,hubert2016learning,li2018heterogeneous,li2019heterogeneous,yao2019heterogeneous,cai2019exploring,khodabandeh2019robust,zhu2019adapting,he2019multi,saito2019strong,kim2019diversify,xie2019multi,hsu2020progressive,zheng2020cross}\\
 & multi-target DA & $N_T>1$ & there are multiple target domains  &  \cite{yu2018multi,gholami2020unsupervised}\\
\hline
\multirow{4}{*}{label set}   & closed-set DA &  $\mathcal{C}_S=\mathcal{C}_T$ & the label sets of source and target domains are the same  &  
\cite{zhuo2017deep,shrivastava2017learning,tzeng2017adversarial,wu2019squeezesegv2,chen2018domain,long2015learning,sun2017correlation,rozantsev2018beyond,zellinger2017central,chen2020homm,kang2019contrastive,long2017deep,li2018adaptive,cariucci2017autodial,lee2019sliced,roy2019unsupervised,hoffman2016fcns,ganin2016domain,chen2017no,hong2018conditional,tsai2018learning,long2018conditional,cicek2019unsupervised,hu2020panda,liu2019transferable,xu2019adversarial,du2020dual,chen2020adversarial,pan2019transferrable,bousmalis2017unsupervised,hoffman2018cycada,kang2018deep,sankaranarayanan2018generate,tran2019gotta,li2019cycle,wu2018dcan,liu2017unsupervised,russo2018source,ghifary2016deep,bousmalis2016domain,sun2019unsupervised,carlucci2019domain,xu2019self,feng2019self,kim2020cross,achituve2020self,tzeng2018splat,zhang2018fully,murez2018image,zhao2019learning,combes2020domain,lipton2018detecting,tan2019generalized,azizzadenesheli2019regularized,zou2018unsupervised,zou2019confidence,laine2017temporal,tarvainen2017mean,french2018self,athiwaratkun2019there,zhang2017curriculum,wang2018visual,saito2018maximum,chen2018road,li2014learning,tzeng2015simultaneous,ballester2019semi,saito2019semi,wang2019weakly,shu2019transferable,duan2009domain,sun2011two,duan2012exploiting,chattopadhyay2012multisource,duan2012domain,yang2007cross,schweikert2009empirical,xu2012multi,sun2013bayesian,bhatt2016multi,xu2018deep,zhao2018adversarial,hoffman2018algorithms,peng2019moment,zhao2019multi,zhao2020distilling,lin2020multi,yu2018multi,gholami2020unsupervised,wang2011heterogeneous,hubert2016learning,li2018heterogeneous,li2019heterogeneous,yao2019heterogeneous,cai2019exploring,khodabandeh2019robust,zhu2019adapting,he2019multi,saito2019strong,kim2019diversify,xie2019multi,hsu2020progressive,zheng2020cross}\\
 & open-set DA & $\mathcal{C}_S \cap \mathcal{C}_T \subset \mathcal{C}_T$ & source label set is a proper subset of target label set & \cite{panareda2017open,saito2018open,liu2019separate} \\
 & partial DA & $\mathcal{C}_S \cap \mathcal{C}_T \subset \mathcal{C}_S$ &  target label set is a proper subset of source label set & \cite{cao2018partial,zhang2018importance,cao2018partialadversarial,cao2019learning} \\
 & universal DA & $\mathcal{C}_S\ ?\ \mathcal{C}_T$ & no prior knowledge of the label sets is available & \cite{you2019universal}\\
\hline
\multirow{2}{*}{target data}   & domain adaptation &  with $\mathbf{X}_T$ & target data is known during training & 
\cite{zhuo2017deep,shrivastava2017learning,tzeng2017adversarial,wu2019squeezesegv2,chen2018domain,long2015learning,sun2017correlation,rozantsev2018beyond,zellinger2017central,chen2020homm,kang2019contrastive,long2017deep,li2018adaptive,cariucci2017autodial,lee2019sliced,roy2019unsupervised,hoffman2016fcns,ganin2016domain,chen2017no,hong2018conditional,tsai2018learning,long2018conditional,cicek2019unsupervised,hu2020panda,liu2019transferable,xu2019adversarial,du2020dual,chen2020adversarial,pan2019transferrable,bousmalis2017unsupervised,hoffman2018cycada,kang2018deep,sankaranarayanan2018generate,tran2019gotta,li2019cycle,wu2018dcan,liu2017unsupervised,russo2018source,ghifary2016deep,bousmalis2016domain,sun2019unsupervised,carlucci2019domain,xu2019self,feng2019self,kim2020cross,achituve2020self,tzeng2018splat,zhang2018fully,murez2018image,zhao2019learning,combes2020domain,lipton2018detecting,tan2019generalized,azizzadenesheli2019regularized,zou2018unsupervised,zou2019confidence,laine2017temporal,tarvainen2017mean,french2018self,athiwaratkun2019there,zhang2017curriculum,wang2018visual,saito2018maximum,chen2018road,panareda2017open,saito2018open,liu2019separate,cao2018partial,zhang2018importance,cao2018partialadversarial,cao2019learning,you2019universal,li2014learning,tzeng2015simultaneous,ballester2019semi,saito2019semi,wang2019weakly,shu2019transferable,wang2011heterogeneous,hubert2016learning,li2018heterogeneous,li2019heterogeneous,yao2019heterogeneous,duan2009domain,sun2011two,duan2012exploiting,chattopadhyay2012multisource,duan2012domain,yang2007cross,schweikert2009empirical,xu2012multi,sun2013bayesian,bhatt2016multi,xu2018deep,zhao2018adversarial,hoffman2018algorithms,peng2019moment,zhao2019multi,zhao2020distilling,lin2020multi,yu2018multi,gholami2020unsupervised,cai2019exploring,khodabandeh2019robust,zhu2019adapting,he2019multi,saito2019strong,kim2019diversify,xie2019multi,hsu2020progressive,zheng2020cross} \\
 & domain generalization & without $\mathbf{X}_T$ & target data is unknown during training  & \cite{muandet2013domain,li2018learning,li2018domain} \\
\bottomrule
\end{tabular}
}
\label{tab:Definition}
`\#sources' and `\#targets' respectively represent the number of source domains and the number of target domains; $N_S$, $N_T$, $N_{SL}$, $N_{TL}$ respectively denote the numbers of source samples, target samples, labeled source samples, and labeled target samples; $d_S$, $d_T$ are the data dimensions of source data and target data respectively; $\mathcal{C}_S$, $\mathcal{C}_T$ respectively represent the label set for the source and target domains; $\mathbf{X}_T$ is the target data without labels.
\end{table*}

\paragraph{Organization of This Survey}
In this survey, we review recent progress on visual domain adaptation, comparing advantages and disadvantages of different approaches in this class, and discussing future directions.

In particular: First, we define different DA strategies in Section~\ref{sec:ProblemDefinition}. Second, we summarize the available datasets for performing DA evaluation focusing on computer vision tasks in Section~\ref{sec:Datasets}. And then, we briefly introduce the theoretical view in Section~\ref{ssec:theory}, summarize and compare the representative approaches on different single-source DUDA categories, including discrepancy-based methods (Section~\ref{ssec:Discrepancy}), adversarial discriminative methods (Section~\ref{ssec:Discriminative}), adversarial generative methods (Section~\ref{ssec:Generative}),  self-supervision-based methods (Section~\ref{ssec:selfsupervision}), and combinations and others (Section~\ref{ssec:Others}), followed by both qualitative and quantitative comparisons of different categories in Section~\ref{ssec:qualitative} and Section~\ref{ssec:quantitative}. Finally, we discuss potential future research directions in Section~\ref{sec:NewPerspectives}.



\section{Domain Adaptation Taxonomy}
\label{sec:ProblemDefinition}

We introduce a standard definition of the variables and models to enable effective comparisons and classification. Let $\textbf{x}$ and $y$\footnote{In this paper we assume $\textbf{x}$ is an image and $y$ could be any label type (\textit{e.g.} object classes, bounding boxes, semantic segmentation, etc).} respectively denote the input data and output label, drawn from a specific domain probability distribution $P(\textbf{x},y)$.
In typical domain adaptation, there is one source domain and one target domain. Suppose the source data and corresponding labels drawn from the source distribution $P_S(\textbf{x},y)$ are $\textbf{X}_S$ and $Y_S$ respectively, and the target data and corresponding labels drawn from the target distribution $P_T(\textbf{x},y)$ are $\textbf{X}_T$ and $Y_T$ respectively. The corresponding marginal distributions include $P_S(\textbf{x})$, $P_S(y)$, $P_T(\textbf{x})$, $P_T(y)$, and conditional distributions include $P_S(\textbf{x}|y)$, $P_S(y|\textbf{x})$, $P_T(\textbf{x}|y)$, $P_T(y|\textbf{x})$. Three typical sources of variation between the two domains considered in the literature include: 
\begin{enumerate}
\item covariate shift, where $P_S(y\mid\textbf{x})=P_T(y\mid\textbf{x})$ for all $\textbf{x}$, but $P_S(\textbf{x})\neq P_T(\textbf{x})$;
\item label shift, where $P_S(\mathbf{x}\mid y) = P_T(\mathbf{x}\mid y)$ for all $y$, but $P_S(y)\neq P_T(y)$;
\item concept drift, where $P_S(\textbf{x},y)\neq P_T(\textbf{x},y)$.
\end{enumerate}
In addition, the source dataset is $D_S=\{\textbf{X}_S,Y_S\}=\{(\textbf{x}_S^i,y_S^i)\}_{i=1}^{N_S}$, the target dataset is $D_T=\{\textbf{X}_T,Y_T\}=\{(\textbf{x}_T^j,y_T^j)\}_{j=1}^{N_T}$, where $N_S$ and $N_T$ are the number of source samples and target samples respectively, $\textbf{x}_S^i\in \mathds{R}^{d_S}$ and $\textbf{x}_T^j\in \mathds{R}^{d_T}$ are referred as observations in the source domain and the target domain, and $y_S^i$ and $y_T^j$ are the corresponding class labels.

Suppose the number of labeled target samples is $N_{TL}$; then, the DA problem can be classified into different categories:
\begin{itemize}
\item \emph{unsupervised DA}, when $N_{TL}=0$;
\item \emph{fully supervised DA}, when $N_{TL}=N_T$;
\item \emph{semi-supervised DA}, otherwise.
\end{itemize}

Suppose the number of labeled source samples is $N_{SL}$; then, DA can be classified into:
\begin{itemize}
\item \emph{strongly supervised DA}, when $N_{SL}=N_S$;
\item \emph{weakly supervised DA},  otherwise.
\end{itemize}

Based on the representations, $d_S$ and $d_T$, of the source and target domains~(\textit{e.g.} images vs. text),  we can classify DA into:
\begin{itemize}
\item \emph{homogeneous DA}, when $d_S=d_T$;
\item \emph{heterogeneous DA}, otherwise.
\end{itemize}

Suppose the number of source domains is $N_S$; then, the DA task can be categorized into:
\begin{itemize}
\item \emph{single-source DA}, when $N_S=1$;
\item \emph{multi-source DA}, when $N_S>1$.
\end{itemize}

Similarly, let $N_T$ denote the number of target domains; we can then classify DA into:
\begin{itemize}
\item \emph{single-target DA}, when $N_T=1$;
\item \emph{multi-target DA}, when $N_T>1$.
\end{itemize}


\begin{table*}[!t]
\centering
\caption{Released and freely available datasets for domain adaptation, where `\# Samples' represents the total number of samples, 'Type' indicates whether the data is collected from simulation or real world, 'syn' is short for 'synthetic', 'cla', 'reg', 'det', and 'seg' are short for 'classification', 'regression', 'detection', and 'segmentation', respectively.}
\resizebox{0.95\textwidth}{!}{%
\begin{tabular}
{cccccccccc}
\toprule
\textbf{Task} & \textbf{Dataset} & \textbf{Ref} & \textbf{\# Samples} & \textbf{Labels} & \textbf{Resolution} & \textbf{Type} & \textbf{Task}& \textbf{Short description}\\
\hline
\multirow{5}{*}{digit recognition} & MNIST (M) & \cite{lecun1998gradient}  & 70K & 10 classes &  $28\times28$  & real & cla  &  size-normalized and centered handwritten digits\\
& MNIST-M (M-M) & \cite{ganin2015unsupervised}  & 149,002 & 10 classes &  32 $\times$ 32  & real & cla &  MNIST digits with color photos as their background\\
& USPS (U) &  \cite{hull1994database} &  9,298 & 10 classes &  16 $\times$ 16  & real  &  cla &  unconstrained handwritten digits\\
& SVHN (S) & \cite{netzer2011reading}  & 99,289  &  10 classes &  32 $\times$ 32  & real  &  cla & digits obtained from house numbers in Google Street View images \\
& Synthetic digits (SD)& \cite{ganin2015unsupervised}  & 500K  &  10 classes &   32 $\times$ 32 & syn  &  cla & digits generated from Windows fonts by varying types of conditions\\
\hline
\multirow{5}{*}{object classification} & Office-31 (O) & \cite{saenko2010adapting}  & 4,110 & 31 classes &  -  & real & cla & images from 3 domains: Amazon (A), Webcam (W) and DSLR (D)\\
& Office+Caltech (OC) &  \cite{gong2013connecting} &  2,533 & 10 classes & - & real  & cla  &  shared categories of Office-31 and Caltech-256 (C)\\
& Office-Home (OH) &  \cite{venkateswara2017deep} &  15,500 & 65 classes & - & real  & cla  &  images from 4 domains: Ar, Cl, Pr and Rw\\
& VisDA-2017 (V) & \cite{peng2017visda} &  280,157 & 12 classes & -  &  syn/real & cla, seg  & about 280K images from synthetic data to real imagery\\
& DomainNet (DN) & \cite{peng2019moment} &  569,010 & 345 classes &  -   &  syn/real & cla  & the largest DA dataset for object classification with 6 domains\\
\hline
\multirow{3}{*}{pose estimation} & UnityEyes (UE) & \cite{shrivastava2017learning}  & 1.2M  & gaze degrees & $640 \times 480$   & syn  & reg  & synthetic images for eye gaze estimation \\
& MPIIGaze (MG) & \cite{zhang2015appearance}  & 214K  &  gaze degrees & $36 \times 60$   & syn  &  reg & real images for eye gaze estimation  \\
& NYU (N) & \cite{tompson2014real}  & 81,008  & hand pose &  $224 \times 224$   & syn/real  &  reg & images from both syn and real domains for hand pose estimation \\
\hline
\multirow{1}{*}{3D point cloud} & KITTI-LiDAR (K-L)& \cite{geiger2012we}  & 10,848  & 3 classes &  $64\times512$   & real  & cla, seg  &  LiDAR point cloud with point-wise labels for autonomous driving\\
\multirow{1}{*}{segmentation} & GTA-LiDAR (G-L) & \cite{wu2019squeezesegv2}  &  100K & 2 classes &  $64\times512$   & syn  &  cla, seg & LiDAR point cloud synthesized in GTA-V \\
\hline
\multirow{5}{*}{object detection} & Cityscapes (CS) & \cite{cordts2016cityscapes}  & 3,475 & 8 classes &  $2048 \times 1024$   & real  & det  & converted from Cityscapes with instance segmentation mask \\

& Foggy Cityscapes~(FC) & \cite{sakaridis2018semantic}&  3,475 & 8 classes &  $2048 \times 1024$   & syn  & det  & adding synthetic for to the original Cityscapes images\\
& SIM10k (SM) & \cite{johnson2017driving}&  10,000 & 8 classes &  $1914 \times 1052$   & syn  & det  & images synthesized in GTA-V for object detection\\
& KITTI-Obj (K-O) & \cite{geiger2012we}&  7,481 & 8 classes &  $1250 \times 375$   & real  & det  & images collected from real urban scenes for object detection\\
& Syn2Real-D (S2R) & \cite{peng2018syn2real} & 248K & 13 classes & - & syn/real & det & images from 3D CAD models and real-world detection datasets\\

\hline
\multirow{4}{*}{semantic segmentation} & Cityscapes (CS) & \cite{cordts2016cityscapes}  & 5,000 & 30 classes &  $2048 \times 1024$   & real  & seg  & images collected from real urban scenes for semantic segmentation \\
& BDDS (B)& \cite{yu2018bdd100k} & 8,000& 19 classes &  $1280 \times 720$  & real & seg & images collected from real urban scenes for semantic segmentation\\
& GTA (G) & \cite{richter2016playing}  & 24,966 &  19 classes &  $1914\times1052$  & syn  & seg  & images synthesized in GTA-V for semantic segmentation\\
& SYNTHIA (SY) & \cite{ros2016synthia}  & 400K  & 16 classes &   $960\times720$  & syn  & seg  & synthetic urban images for semantic segmentation\\ 
\bottomrule
\end{tabular}
}
\label{tab:Dataset}
\end{table*}

Let $\mathcal{C}_S$ and $\mathcal{C}_T$ respectively denote the label set for the source and target domains; then, we can define DA into the following different categories:
\begin{itemize}
\item \emph{closed set DA}, when $\mathcal{C}_S=\mathcal{C}_T$;
\item \emph{open set DA}, when $\mathcal{C}_S \cap \mathcal{C}_T \subset \mathcal{C}_T$;
\item \emph{partial DA}, when $\mathcal{C}_S \cap \mathcal{C}_T \subset \mathcal{C}_S$;
\item \emph{universal DA}, when no prior knowledge of the label sets is available;
\end{itemize}
where $\cap$ and $\subset$ indicate intersection and proper subset.

Although without labels, the target data is usually available during training in typical DA. If the target data is also unavailable, we often denote this task as domain generalization or zero-shot DA. Therefore, we have:
\begin{itemize}
\item \emph{domain adaptation}, when $\textbf{X}_T$ is available during training;
\item \emph{domain generalization or zero-shot DA}, when $\textbf{X}_T$ is unavailable during training.
\end{itemize}

The classification of different DA categories are summarized in Table~\ref{tab:Definition}. We focus on the review of recent unsupervised domain adaptation (UDA) methods under homogeneous, single-source, single-target, strongly-supervised, and closed-set settings, \textit{i.e.} $N_{TL}=0$, $d_S=d_T$, $N_S=1$, $N_T=1$, $N_{SL}=N_S$, $\mathcal{C}_S=\mathcal{C}_T$. The goal is to learn a model $f$ that can correctly predict a sample from the target domain based on labeled $\{\textbf{X}_{S},Y_{S}\}$ and unlabeled $\{\textbf{X}_T\}$.


\section{Datasets}
\label{sec:Datasets}
In the early years, the datasets for DA were mainly collected from real world scenarios. Increasingly, large-scale synthetic datasets are being generated from simulation engines with labels automatically obtained, which induces large domain shift from the real world data.  The released datasets are summarized in Table~\ref{tab:Dataset}.

\textbf{Digit recognition.}
MNIST~\cite{lecun1998gradient} is a dataset of handwritten digits with a training set of 60K examples and a test set of 10K examples. The digits have been size-normalized and centered in a fixed-size image. USPS~\cite{hull1994database} is also a dataset of handwritten digits with 7,291 examples for training and 2,007 examples for testing. 
MNIST-M~\cite{ganin2015unsupervised} is created by combining MNIST digits with the patches randomly extracted from color photos of BSDS500~\cite{arbelaez2011contour} as their background. It contains 59,001 training and 90,001 test images. Synthetic Digits \cite{ganin2015unsupervised} consists of 500K images generated from Windows fonts by varying the text (that includes one-, two-, and three-digit numbers), positioning, orientation, background and stroke colors, and the amount of blur.
SVHN \cite{netzer2011reading} is obtained from house numbers in Google Street View images. It contains 73,257 digits for training, and 26,032 digits for testing. 


\textbf{Object classification.} Office-31~\cite{saenko2010adapting} is a standard benchmark for DA. There are 4,110 images within 31 categories collected from office environments in 3 image domains: Amazon (A) with 2,817 images downloaded from amazon.com, Webcam (W), and DSLR (D), with 795 and 498 images taken by web camera and digital SLR camera with different photographical settings, respectively.

Office+Caltech~\cite{gong2013connecting} consists of the 10 overlapping categories shared by Office-31~\cite{saenko2010adapting} and Caltech-256 (C)~\cite{griffin2007caltech}.

Office-Home~\cite{venkateswara2017deep} consists of about 15,500 images from 65 categories of everyday objects in office and home settings. There are 4 different domains: Artistic images (Ar), Clip Art (Cl), Product images (Pr) and Real-World images (Rw).

VisDA-2017~\cite{peng2017visda} is a challenging testbed for UDA with the domain shift from synthetic data to real imagery. There are about 280K images from 12 categories, including a training set with 152,397 synthetic images, a validation set with 55,388 real-world images, and a test set with 72,372 real-world images. The dataset is used in the Visual Domain Adaptation Challenge, including classification and segmentation tracks.


DomainNet~\cite{peng2019moment}, the largest DA dataset to date for object classification, containing about 600K images from 6 domains: Clipart, Infograph, Painting, Quickdraw, Real, and Sketch. There are 345 object categories altogether.

\textbf{Pose estimation.} UnityEyes~\cite{shrivastava2017learning} is a synthetic dataset with 1.2M images for eye gaze estimation. MPIIGaze~\cite{zhang2015appearance} contains 214K real eye gaze images captured under extreme illumination conditions.

NYU~\cite{tompson2014real} is a hand pose dataset with 72,757 training frames and 8,251 testing frames captured by 3 in 1 frontal and 2 side views. Each depth frame is labeled with hand pose information that is used to create a synthetic depth image.


\textbf{3D point cloud segmentation.} KITTI~\cite{geiger2012we} is a real-world dataset for autonomous driving with images, LiDAR scans, and 3D bounding boxes organized in sequences. KITTI-LiDAR consists of 10,848 samples with point-wise labels obtained from 3D bounding boxes. Each point belongs to a car, pedestrian, or cyclist. 8,057 samples are used for training and the rest 2,791 samples are for testing.


GTA-LiDAR~\cite{wu2019squeezesegv2} contains 100K LiDAR point clouds synthesized in GTA-V using the method in~\cite{yue2018lidar} to do Image-LiDAR registration. The wide variety of scenes, car types, and traffic conditions, which ensures the diversity of the synthetic data. The categories car and pedestrian are synthesized.

\textbf{Object detection.} Cityscapes~\cite{cordts2016cityscapes} is a dataset of real urban scenes containing 3,475 images with pixel-level annotations. Since it is not designed for object detection, tightest rectangle of an instance segmentation mask is used as the ground truth bounding box~\cite{chen2018domain}. 

Foggy Cityscapes~\cite{sakaridis2018semantic} is a recently proposed synthetic foggy dataset simulating fog on real scenes. The depth maps provided in Cityscapes are used to simulate three intensity levels of fog~\cite{sakaridis2018semantic}. Each foggy image is synthesized from an image with depth map from Cityscapes.

SIM10k~\cite{johnson2017driving} is a synthetic dataset collected from the computer game GTA-V. It contains 10,000 images with bounding box annotations for cars. 

KITTI-Obj~\cite{geiger2012we} is a real-world dataset consisting of 7,481 labeled images. The images are collected from various traffic situations, including freeways, rural and urban areas.

Syn2Real-D~\cite{peng2018syn2real} is a dataset consisting of 248K images. The synthetic images are collected from 3D CAD models while the real-world images are from COCO~\cite{lin2014microsoft} and YouTube Bounding Boxes~\cite{real2017youtube}.

\textbf{Semantic segmentation.} Cityscapes~\cite{cordts2016cityscapes} contains vehicle-centric urban street images collected from a moving vehicle in 50 cities from Germany and neighboring countries. There are 5,000 images with pixel-wise annotations, including a training set with 2,975 images, a  validation set with 500 images, and a test set with 1,595 images. It is widely used in segmentation.


BDDS~\cite{yu2018bdd100k} contains thousands of real-world dashcam video frames with accurate pixel-wise annotations.  It has a  label space that is compatible with Cityscapes.

GTA~\cite{richter2016playing} is collected in the high-fidelity rendered computer game GTA-V with pixel-wise semantic labels. It contains 24,966 images (video frames). There are 19 classes that are compatible with Cityscapes.

SYNTHIA~\cite{ros2016synthia} is a large synthetic dataset. To pair with Cityscapes, a subset, named SYNTHIA-RANDCITYSCAPES, was designed with 9,400 images which are automatically annotated with 16 compatible classes, one void class, and some unnamed classes.

\section{Single-Source DUDA}
\label{sec:SingleSource}

In this section, we first introduce a theoretical view for domain adaptation. Second, we summarize different categories of single-source DUDA. Finally, we compare the advantages and disadvantages of these methods.

\subsection{Theory Brief}
\label{ssec:theory}

Many methods in the domain of machine learning are based on empirical evidence rather than well-founded theory. The ones that have solid theory background use statistics profusely. Domain adaptation is no exception. However, 
upper bounds on the generalization target error by learning from the source data have been obtained. In a seminal paper, \citet{ben2010theory} provided a bound for domain adaptation on the target risk that generalizes the standard bound on the source risk. Informally, the theory says that if there exists a common hypothesis (classifier) that generalizes well on both the source and the target domains, the performance difference of any classifier on these two domains could be bounded by the distance between the data distributions of the two domains. The authors proposed $\HH$-divergence, a parametric pseudo-metric to measure the distance between two domains.  Using this this pseudo-metric, two domains are considered close if there exists a binary classifier (a discriminator) that, upon seeing data from the two domains, can distinguish which domain the data comes from. This work formalizes the intuitive notion that reducing the two distributions while ensuring a low error on the source domain, yields accurate results in the target domain.  Further, the theory justifies the basis of many recent DA algorithms that learn domain-invariant representations, using either domain adversarial classifier or discrepancy-based approaches. 

\citet{mansour2009domain} introduced a new divergence measure: the discrepancy distance, which was used to provide a generalization guarantee for the target domain. Compared with the $\HH$-divergence that can only be used in the setting of binary classification, the discrepancy distance could be used to achieve a generalization bound for target domain in regression setting as well. In a later work,~\citet{mansour2014robust} derived generalization bounds on the target error by making use of the robustness properties introduced in~\citep{xu2012robustness}. Extensions of the above theory to multi-source domain adaptation for both classification and regression problems also exist~\citep{mansour2009multiple,zhao2018adversarial}.

\subsection{Discrepancy-based Methods (Table~\ref{tab:Discrepancy})}
\label{ssec:Discrepancy}

Discrepancy-based methods explicitly measure the discrepancy between the source and target domains on corresponding activation layers of the two network streams. Long et al. \cite{long2015learning} designed a Deep Adaptation Network, where the discrepancy is defined as the sum of the multiple kernel variant of maximum mean discrepancies (MK-MMD) between the fully connected (FC) layers.
Sun et al. \cite{sun2017correlation} proposed correlation alignment (CORAL) to minimize domain shift by aligning the second-order statistics of the source and target features of the last FL layer. Apart from the CORAL loss on the last FL layer, Zhuo et al. \cite{zhuo2017deep} also incorporated the CORAL loss on the last convolutional (conv) layer. To deal with the high dimension of convolutional layer activations, activation-based attention mapping is employed to distill it into low dimensional representations. The CORAL losses on both the last convolutional layer and the last FC layer are combined.
Wu et al. \cite{wu2019squeezesegv2} studied the UDA problem for 3D LiDAR point cloud segmentation from synthetic data to real world data. At every batch of training, in addition to the focal loss to learn semantics from the point cloud on the synthetic batch, they employed the geodesic distance to penalize discrepancies between batch statistics from two domains. 
In recent papers, Zellinger et al. \cite{zellinger2017central} proposed to match the higher order central moments of probability distributions by means of order-wise moment differences. They utilized the equivalent representation of probability distributions by moment sequences to define a new distance function, called Central Moment Discrepancy (CMD). Chen et al. \cite{chen2020homm} explored the benefits of using higher-order statistics (in this case mainly third-order and fourth-order statistics) for domain matching. They proposed a Higher-order Moment Matching (HoMM) method, and further extended the HoMM into reproducing kernel Hilbert spaces (RKHS). 
Some other types of divergence are also designed to align the source and target domains.
Lee et al. \cite{lee2019sliced} designed sliced Wasserstein discrepancy (SWD) to capture the natural notion of dissimilarity between the outputs of task-specific classifiers. It provides a geometrically meaningful guidance to detect target samples that are far from the support of the source and enables efficient distribution alignment in an end-to-end trainable fashion.
Roy et al. \cite{roy2019unsupervised} proposed domain alignment layers which implement feature whitening for the purpose of matching source and target feature distributions. Additionally, they leveraged the unlabeled target data by proposing the Min-Entropy Consensus loss, which regularizes training while avoiding the adoption of many user-defined hyper-parameters.

\begin{table}[!t]
\centering\scriptsize
\caption{Comparison of different discrepancy-based methods, where `Discrepancy' indicates the discrepancy loss, `Loss level' indicates what level the loss is applied to, `Layer' represents the layers that the loss functions on, `Weight' indicates whether the weights of the two networks are shared or not, `Distribution' indicates what type of distribution is aligned.}
\begin{tabular}
{c | c c c c c c c c }
\toprule
\textbf{Ref} & \textbf{Discrepancy} & \textbf{Loss level} & \textbf{Layer} & \textbf{Weight} & \textbf{Distribution}\\
\hline 
\cite{long2015learning} & MK-MMD & domain & FL & shared & marginal\\
\cite{sun2017correlation} & CORAL & domain & last FL & shared & marginal\\
\cite{zhuo2017deep} & CORAL & domain & last (conv, FL) & shared & marginal\\
\cite{rozantsev2018beyond} & weight, MMD & domain & all & linear & marginal\\
\cite{zellinger2017central} & CMD & domain & all & shared & marginal\\
\cite{chen2020homm} & HoMM & domain & FL & shared & marginal\\
\cite{kang2019contrastive} & CDD & class & all except BN & shared & marginal\\
\cite{pan2019transferrable} & RKHS & class & all & shared & marginal\\
\cite{wu2019squeezesegv2} & Geodesic & domain & all & shared & marginal\\
\cite{long2017deep} & JMMD & domain & FL & shared & joint\\
\cite{li2018adaptive} & Implicit & domain & BN & shared & marginal\\
\cite{cariucci2017autodial} & Implicit & domain & weighted BN & shared & marginal\\
\cite{lee2019sliced} & SWD & domain & all & shared & marginal\\
\cite{roy2019unsupervised} & MEC & domain & DWT layer & shared & marginal\\
\bottomrule
\end{tabular}
\label{tab:Discrepancy}
\end{table}

\begin{table*}[!t]
\centering\small
\caption{Comparison of different adversarial discriminative models, where `En' is short for encoder. Adversarial level refers to the level of alignment for the discriminator input, either globally or class-wisely. }
\resizebox{0.9\textwidth}{!}{
\begin{tabular}
{c | c c c c c c}
\toprule
\textbf{Ref} & \textbf{Adversarial level} &\textbf{En weight} & \textbf{Discriminator input} & \textbf{Discriminator type} & \textbf{Discriminator loss}\\
\hline
\cite{hoffman2016fcns} & global & shared & $F_s, F_t$ &  feature discriminator & GAN loss\\
\cite{ganin2016domain} & global & shared & $F_s, F_t$ & gradient reversal layers & Cross-Entropy\\
\cite{tzeng2017adversarial} & global & unshared & $F_s, F_t, \tilde{y}_s, \tilde{y}_t$ & feature, output discriminator &GAN loss \\
\cite{chen2017no} & class, global & shared & $F_s, F_t, \tilde{y}_s, \tilde{y}_t$ & feature discriminator & GAN loss\\
\cite{hong2018conditional} & global & shared & $(F_s + F_{noise}), F_t$ & feature discriminator & GAN loss\\
\cite{tsai2018learning} & global & shared & $\tilde{y}_s, \tilde{y}_t$, & output discriminator &GAN loss\\
\cite{long2018conditional} & global & shared & $F_s, F_t, \tilde{y}_s, \tilde{y}_t$ & conditional discriminator & GAN loss, Conditional Entropy \\
\cite{cicek2019unsupervised} & global & shared & $i_{domain}$, $\tilde{y}_s, \tilde{y}_t$ & joint discriminator & Cross-Entropy & \\
\cite{hu2020panda} & global & unshared & $F_s, F_t, \hat{F}_{s}, \hat{F}_{t}$ &feature, prototypical discriminator & GAN loss & \\
\cite{liu2019transferable} & global & shared & $F_s, F_t $ & feature discriminator & GAN loss & \\
\cite{xu2019adversarial} & global & shared & $F_s, F_t, F_m, \mathbf{x}_s, \mathbf{x}_t, \mathbf{x}_m$ & output discriminator & GAN loss & \\
\cite{du2020dual} & class & shared & $F_{s}, F_{t}$ & joint discriminator & GAN loss \\
\cite{chen2020adversarial} & global & shared & $\mathbf{x}_s, \mathbf{x}_t$ & gradient reversal layers & Binary Cross-Entropy of confusion matrix & \\
\cite{chen2018domain} & global, instance & shared & $F_s, F_t$ & feature discriminator & GAN loss\\
\cite{xie2019multi} & global & shared & $F_s, F_t$ & feature discriminator & GAN loss\\
\cite{zheng2020cross} & global & shared & $F_s, F_t$ & gradient reversal layers & Cross-Entropy \\
\cite{zhu2019adapting} & global & shared & $F_s, F_t$ & feature discriminators &GAN loss\\
\cite{he2019multi} & global & shared &$F_s, F_t$ & gradient reversal layers & Cross Entropy \\
\cite{saito2019strong} & global & shared & $F_s, F_t$ & gradient reversal layers & Cross Entropy\\

\bottomrule
\end{tabular}}
\label{tab:Discriminative}
\small{$F_s, F_t, F_{noise}$ : Extracted features of source image, target image, or input noise; $F_m$ : mixed feature of $F_s$ and $F_t$; $\Tilde{y}_s, \Tilde{y}_t$ : Predicted labels of source or target images; $i_{domain}$ : index of domain; $\mathbf{x}_s, \mathbf{x}_t$: source and target images; $\mathbf{x}_m$ : mixed image of $\mathbf{x}_s$ and $\mathbf{x}_t$}.
\end{table*}

Instead of explicitly modeling the discrepancy between the source and the target domains, some papers implicitly minimize domain discrepancy by aligning the Batch Normalization (BN) statistics. Li et al. \cite{li2018adaptive} proposed to adopt domain specific normalization for different domains. The proposed Adaptive BN (AdaBN) replaces the moving average mean and variance of all BN layers in the task network trained on the source domain with the mean and variance estimated from the target mini-batches. AdaBN~\cite{li2018adaptive} and other DUDA methods define a prior on which layers should be adapted. Instead, Cariucci et al. \cite{cariucci2017autodial} proposed to learn automatically which layers of the network should be aligned and the corresponding alignment degree. The Auto-DomaIn Alignment Layer (AutoDIAL) is embedded multiple times to align the learned feature representations of the source and target domains at different levels. These BN-based methods have fewer parameters to tune, higher computational efficiency, and competitive performance.

The methods described above measure the domain discrepancy at the domain level, which neglects the information concerning the classes from which the samples are drawn and thus may lead to misalignment and poor performance. Kang et al. \cite{kang2019contrastive} proposed Contrastive Adaptation Network, which optimizes a new metric, Contrastive Domain Discrepancy (CDD), by explicitly minimizing the intra-class discrepancy and maximizing the inter-class domain discrepancy. The source and target samples of the same underlying class are drawn closer, while the samples from different classes are pushed apart. Pan et al. \cite{pan2019transferrable} recently proposed Transferrable Prototypical Networks, which perform domain alignment such that prototypes for each class in the source and target domains are close in the embedding space and the predictions from prototypes separately on source and target data are similar.

Most of the papers mentioned above consider aligning the marginal distributions in the feature space. When confronted with complex tasks, these approaches would fail when the label distributions are drastically different between source and target domains. The joint alignment of distributions $D_S = (\mathbf{X}_S, Y_S)$ and $D_T = (\mathbf{X}_T, Y_T)$ is considered in~\cite{long2017deep} under the assumptions that $P_S(\mathbf{x}) \neq P_T(\mathbf{x})$ and $P_S(y | \mathbf{x}) \neq P_T(y | \mathbf{x})$. The joint distributions across domains is projected to a Reproducing Kernel Hilbert Space (RKHS) $\mathcal{H}$ and MMD is used as the distance metric. During the joint distribution alignment, the distribution shift $P_S(\mathbf{x})$ and $P_T(\mathbf{x})$, $P_S(y | \mathbf{x})$ and $ P_T(y | \mathbf{x})$ are significantly reduced.

The above methods all adopt weight-sharing between the two streams of the Siamese architecture~\cite{bromley1994signature,koch2015siamese} that attempts to reduce the impact of domain shift by learning domain-invariant features. However, domain invariance may be detrimental to discriminative power. On the contrary, Rozantsev et al. \cite{rozantsev2018beyond} proposed to explicitly model the domain shift and relaxed the weight-sharing constraint to a linear correlation. They jointly optimized a weight regularizer, representing the loss between corresponding layers of the two streams, and an unsupervised regularizer, encoding the MMD measure and favoring similar distributions of the source and target representations.

\subsection{Adversarial Discriminative Models (Table~\ref{tab:Discriminative})}
\label{ssec:Discriminative}


Adversarial discriminative models usually employ an adversarial objective with respect to a domain discriminator to encourage domain confusion (see Table~\ref{tab:Discriminative}). In the early-stage of adversarial discriminative models, the domain adversarial training of neural networks is proposed to learn domain invariant and task discriminative representations~\cite{ganin2016domain}. It is directly derived from the seminal theoretical works of Ben et al. \cite{ben2010theory} and directly optimizes the $\mathcal{H}$-divergence between source and target. By deriving the generalization bound on the target risk and obtaining an empirical formulation of the objective, Ganin et al. \cite{ganin2016domain} proposed the Domain-Adversarial Neural Networks (DANN) algorithm. From this point of view, the adversarial discriminative models are originally similar to the discrepancy-based models. Recently, a couple of adversarial discriminative models were proposed with different algorithms and network architectures, thus differing from the discrepancy-based methods.

Suppose $m_S$ and $m_T$ are the representation mappings of the source and target domains, respectively, and $d$ is a domain discriminator, which classifies whether a data point is drawn from the source or the target domain. The adversarial discriminator is trained typically based on an adversarial loss $\mathcal{L}_{a_d}$. The loss $\mathcal{L}_{a_m}$ used to train representation mapping is different in existing methods. The Domain-Adversarial Neural Network~\cite{ganin2016domain} optimizes the mapping to minimize the discriminator loss directly $\mathcal{L}_{a_m}=-\mathcal{L}_{a_d}$, which might be problematic, since the discriminator converges quickly during training, causing the gradients to vanish. A \textit{gradient reversal} layer was proposed to achieve domain adversarial training with a single feed-forward network with standard backpropagation and stochastic gradient descent. Tzeng et al. \cite{tzeng2017adversarial} proposed Adversarial Discriminative Domain Adaptation~(ADDA), using an inverted label GAN loss to split the optimization process into two independent objectives for the generator and discriminator.

Besides aligning marginal distributions, several methods also align conditional or joint distributions. Long et al. \cite{long2018conditional} considered aligning conditional distribution across domains, and proposed Conditional Domain Adversarial Network (CDAN). Based on DANN~\cite{ganin2016domain}, they used conditional discriminator $\mathbf{D}(\mathbf{f}\times\mathbf{g})$ with improved discriminability, where $\mathbf{f}$ is feature extractor and $\mathbf{g}$ is classifier, to capture the cross-covariance between feature representations and classifier predictions. To extend joint distribution alignment, Du et al. \cite{du2020dual} used dual adversarial strategy to train a dual-discriminator to pit against each other. Cicek et al. \cite{cicek2019unsupervised} also aimed for joint distribution $P(d, y)$ alignment over domain $d$ and label $y$ by a joint predictor and aligned its output with classifier's prediction.After analyzing the drawbacks of feature-level alignment methods, Liu et al. \cite{liu2019transferable} proposed Transferable Adversarial Training (TAT), not only adapting feature representations from difference domains, but also generating transferable examples to make the classifier learn a more robust decision boundary.

Xu et al. \cite{xu2019adversarial} explored two common limitations in current adversarial-based methods. Sampling from source and target domains separately is insufficient to ensure domain-invariance at the whole latent space, and does not give the discriminator a hard label to judge real and fake samples. They proposed a mixed version of the discriminator to guarantee domain-invariance in a more continuous latent space, thus improving the robustness of models performance. Chen et al. \cite{chen2020adversarial} adopted the concept of self-training. They analyzed the noise of pseudo-labels in the confusion matrix and proposed correspondingly an adversarial-learned loss to accurately estimate the confusion matrix. In this way, their proposed method inherits the strength of both adversarial learning and self-training paradigm.

Hoffman et al. \cite{hoffman2016fcns} made the very first effort for domain adaptation in semantic segmentation. They employed a pixel-level adversarial loss to enforce the network to extract domain-invariant features for semantic segmentation and further applied category-specific constraints, \textit{e.g.} pixel percentage histograms. Instead of only performing domain adversarial globally, Chen et al. \cite{chen2017no} proposed to perform feature alignment jointly at the global and class-wise levels by leveraging soft labels from source and target-domain data.  Hong et al. \cite{hong2018conditional} proposed to learn a conditional generator to transform features of synthetic images to real-image like features, and perform domain adversarial training on the learned features. However, the proposed method is network-specific and only applied to the FCN model structure. 
While  previous works mostly perform feature alignment in the middle of a network, Tsai et al. \cite{tsai2018learning} adopted adversarial learning in the output space. To further enhance the adapted model, they constructed a multi-level adversarial network to effectively perform output space domain adaptation at different feature levels. To address DA in object detection, ~\cite{he2019multi, xie2019multi} applied multi-level domain alignment with adversarial training, and Chen et al. ~\cite{chen2018domain} performed domain alignment on both image level and instance level. Weak alignment model was introduce in \cite{saito2019strong} which focused the adversarial alignment loss on images that are globally similar, putting less emphasis on aligning images which are globally dissimilar. Zhu et al. \cite{zhu2019adapting} instead proposed to perform adversarial learning on region level for domain alignment. Recently, Zhent et al. \cite{zheng2020cross} proposed a coarase-to-fine feature adaptation approach for object detection. Different from image level or instance level feature alignment, foreground regions are extracted by attention mechanism, and aligned through multi-layer adversarial learning.
Based on prototypical representations, Hu et al. \cite{hu2020panda} recently proposed a Prototypical Adversarial Learning scheme to align both feature representations and intermediate prototypes across domains.

\subsection{Adversarial Generative Models (Table~\ref{tab:Generative})}
\label{ssec:Generative}

\begin{table*}[!t]
\centering
\caption{Comparison of different adversarial generative models. `Weight' indicates whether the weights of different GANs are shared.}
\resizebox{0.95\textwidth}{!}{%
\begin{tabular}
{c | c c c c c c}
\toprule
\textbf{Ref} & \textbf{Architecture \& loss} & \textbf{Input of GAN} & \textbf{Weight} & \textbf{Generative level} & \textbf{Discriminative level} & \textbf{Specific objective} \\
\hline
\cite{liu2016coupled} & CoGAN & $\textbf{z}$ &partially shared & pixel-level & pixel-level & joint distribution learning without paired images \\
\cite{shrivastava2017learning} & GAN with new $\mathcal{L}_{g}$ & $\textbf{x}_S$ & - & pixel-level & pixel-level  & self-regularization\\
\cite{bousmalis2017unsupervised} & GAN, masked-PMSE & $\textbf{z},\textbf{x}_S$ & - & pixel-level  & pixel-level  & masked-PMSE minimization \\
\cite{hoffman2018cycada} & CycleGAN, semantic, feature & $\textbf{x}_S,\textbf{x}_T, fea$ & unshared & pixel-level & pixel-level, feature-level & semantic consistency enforcement\\
\cite{kang2018deep} & CycleGAN, attention map & $\textbf{x}_S,\textbf{x}_T$ & unshared & pixel-level & pixel-level, feature-level & adversarial attention alignment \\
\cite{sankaranarayanan2018generate} & GAN with class labels supervision & $\textbf{x}_S,\textbf{x}_T$ & shared & pixel-level & pixel-level & class-consistent generation \\
\cite{tran2019gotta} & CycleGAN with class labels supervision & $\textbf{x}_S,\textbf{x}_T$ & shared & pixel-level & pixel-level, feature-level & attribute-conditioned, photometric transformation \\
\cite{li2019cycle} & CycleGAN with conditional loss & $\textbf{x}_S,\textbf{x}_T$ & shared & pixel-level & pixel-level, feature-level & conditioned on classifier prediction \\
\cite{wu2018dcan} & CycleGAN with $L_{c}$ & $\textbf{x}_S,\textbf{x}_T$ & shared & pixel-level & pixel-level & channel-wise statistics feature alignment \\
\cite{liu2017unsupervised} & Coupled GAN, VAE, cycle-consisteny & $\textbf{x}_S,\textbf{x}_T$ & partially shared & pixel-level & pixel-level & joint-distribution alignment in latent space \\
\cite{kim2019diversify}& GAN, constraint loss $L_{con}$ & $\textbf{x}_S,\textbf{x}_T$ & shared & pixel-level & pixel-level, feature-level & domain diversification \\
\cite{hsu2020progressive} & CycleGAN, feature & $\textbf{x}_S,\textbf{x}_T$ & shared & pixel-level & feature-level & pixel-level and feature-level alignment \\
\bottomrule
\end{tabular}}
\label{tab:Generative}
\end{table*}

Adversarial generative models combine the domain discriminative model with a generative component generally based on generative adversarial nets (GANs)~\cite{goodfellow2014generative}, which include a generator $g$ and a discriminator $d$. $g$ takes random noise $\textbf{z}$ as input to generate a virtual image, while $d$ takes the output of $g$ and real images $\textbf{x}$ as input to classify whether an image is real or generated.  The learning process is that $d$ tries to maximize the probability of correctly classifying real and generated images, while $g$ tries to generate images to maximize the probability of $d$ making a mistake. The Coupled Generative Adversarial Networks (CoGAN)~\cite{liu2016coupled} is composed of a tuple of GANs, and each is responsible for synthesizing images in one domain. CoGAN corresponds to a constrained min-max game of two teams, each with two players.

Instead of taking random noise as input, the generator of more recent GAN based methods is usually conditioned on the source data. Shrivastava et al. \cite{shrivastava2017learning} proposed simulated and unsupervised learning (SimGAN) to improve the realism of a simulator's output using unlabeled real data. The discriminator's loss in SimGAN is the same as is used in a traditional GAN, while a self-regularization loss is added in the refiner (generator) loss to ensure that the refined data do not change much, which aims to preserve the annotation information. The generator in pixel-level DA~\cite{bousmalis2017unsupervised} is conditioned on both a noise vector and an image from the source domain. To penalize large low-level differences between the source and generated images for foreground pixels only, the model learns to minimize a masked Pairwise Mean Squared Error (PMSE) which only calculates the masked pixels (foreground) of the source and the generated images.
Sankaranarayanan et al. \cite{sankaranarayanan2018generate} proposed to learn a mutual feature embedding for source and target images, and to generate intermediate domain images from source and target embeddings. They also designed a multi-class discriminator to encourage the model to extract more class-discriminative features. 

To overcome the under-constrained nature of GAN, ~\cite{zhu2017unpaired} proposed CycleGAN with a cycle-consistency constraint. Based on the CycleGAN loss, some effective adaptation methods were introduced.  Hoffman et al. \cite{hoffman2018cycada} proposed discriminatively-trained Cycle-Consistent Adversarial Domain Adaptation (CyCADA), which adapts representations at both the pixel-level and feature-level, enforces cycle-consistency, and leverages a task loss to perform semantic segmentation adaptation.
Similarly, Russo et al. \cite{russo2018source} introduced bi-directional image translation mapping and proposed class-consistency loss. While CycleGAN~\cite{zhu2017unpaired} can only translate low-level appearance, \textit{e.g.} texture, \cite{tran2019gotta} realized multiple view-point transformation combining with key-point detection network. Similarly, Tzeng et al. \cite{tzeng2018splat} performed domain adaptation on object detection using pixel-level alignment and feature-level alignment.
Extending previous CycleGAN-based works~\cite{zhu2017unpaired,hoffman2018cycada, tzeng2018splat}, Li et al. ~\cite{li2019cycle} proposed cycle-consistent conditional adversarial transfer networks (3CATN) to improve adversarial training and feature generation process by conditioning on the classifier prediction.
Instead of using a discriminator, Wu et al. \cite{wu2018dcan} explored channel-wise statistics alignment of CNN features to guide the generation process.
Liu et al. \cite{liu2017unsupervised} combined CoGAN~\cite{liu2016coupled} with Variational Autoencoder (VAE)~\cite{kingma2013auto} to perform unsupervised image-to-image translation. 
A shared latent space between source and target domains is inferred to align the joint distributions of different domains. And then training data closer to the target domain can be sampled from the shared latent space.
Besides the CycleGAN loss, Kang et al. \cite{kang2018deep} proposed to impose the attention alignment penalty to reduce the discrepancy of attention maps across domains. To make the attention mechanism invariant to domain shift, the target network is trained with a mixture of real and synthetic data from both source and target domains. 
Hsu et al. \cite{hsu2020progressive} leveraged CycleGAN together with feature-level alignment for object detection adaptation. 
Kim et al. \cite{kim2019diversify} further proposed to generate diversified intermediate domains to help domain-invariant representation learning for object detection. A multi-domain discriminator is leveraged to encourage the feature to be indistinguishable among the domains.

\subsection{Self-supervision-based Methods (Table~\ref{tab:Reconstruction})}
\label{ssec:selfsupervision}

\begin{table*}[!t]
\centering
\caption{Comparison of different self-supervision-based methods, where `En' and `De' are short for encoder and decoder, `$N_{En}$' and `$N_{De}$' respectively indicate the number of encoders and decoders, `Loss' indicates the employed self-supervision loss, `S' and `T' in the `Domain' column represent source domain and target domain in which the reconstruction is performed.}
\resizebox{0.9\textwidth}{!}{%
\begin{tabular}
{c | c c c c c c c c c c c}
\toprule
\textbf{Ref} & \textbf{$N_{En}$} & \textbf{En base net} & \textbf{$N_{De}$} & \textbf{Domain} & \textbf{De weight} & \textbf{Self-supervision tasks} & \textbf{Loss} \\
\hline
\cite{ghifary2015domain} & 1  & shared &  1 & S, T  & unshared  & Reconstruction&  L2 \\
\cite{ghifary2016deep} & 1  & shared &  1 & T  & --  & Reconstruction&  L2 \\
\cite{bousmalis2016domain} &  2 & shared/unshared & 1  &  S, T & shared   &Reconstruction & SIMSE\\
\cite{sun2019unsupervised} & 1 & shared  & 3 &  S, T & shared  & Image rotation, Patch Location, Flip prediction & Cross-Entropy (CE) \\
\cite{carlucci2019domain} & 1 & shared  & 1 &  S, T & shared  & Jigsaw Puzzle & CE \\
\cite{xu2019self} & 1 & shared  & 1 &  T & --  & Image rotation, Spatial-aware rotation prediction & CE \\
\cite{feng2019self} & 1 & shared & 1 & S, T & shared & MI minimization \& maximization & Mutual Information~(MI) \\
\cite{kim2020cross} & 1 &shared & 0 & S, T & --  & Instance Discrimination \& Cross-domain entropy minimization & CE \& Entropy\\
\cite{achituve2020self} & 1 & shared & 1 & S, T & shared & Region Reconstruction& L2 \\
\bottomrule
\end{tabular}
}
\label{tab:Reconstruction}
\end{table*}

Self-supervision based methods incorporate auxiliary self-supervised learning task(s) into the original task network. Training the self-supervision task jointly with the original task network is helpful to bring the source and target domains closer. 
Ghifary et al. \cite{ghifary2015domain} designed a three-layer Multi-task Autoencoder (MTAE) architecture to transform the original image into analogs in multiple related domains.
The hidden-input and hidden-output weights represent shared and domain-specific parameters, respectively. 
The learned features are then used as input to a classifier.
The category-level correspondence across domains is required. Self-domain and between-domain reconstruction tasks are introduced as the self-supervision task and are performed during training. Deep reconstruction classification network (DRCN) \cite{ghifary2016deep} combines a convolutional supervised network for source label prediction with a de-convolutional unsupervised network for target data reconstruction. The feature mapping parameters of the two streams are shared, while the labeling parameters of the supervised network and the decoding parameters of the unsupervised network are learned individually.

MTAE requires that the number of samples of corresponding categories in the two domains should be the same. After the sample selection procedure, some important information may be missing. Further, the output of the algorithm is learned features, based on which a classifier (multi-class Support Vector Machine with a linear kernel in this paper) needs to be trained. DRCN employs an end-to-end strategy, without the requirement of aligned pairs. The above two methods use the same encoder to extract domain-invariant features, ignoring the individual characteristics of each domain. Bousmalis et al. \cite{bousmalis2016domain} explicitly learned to extract image representations that are partitioned into two subspaces. One component is private to each domain, which aims to capture domain-specific properties, such as background. The other is shared across domains with the goal of capturing shared representations by using autoencoders and explicit loss functions, \textit{i.e.} scale-invariant mean-square error (SIMSE).

Except for the reconstruction task, 
more recent self-supervision tasks (\textit{e.g.} image rotation prediction and jigsaw prediction) have been used for DA~\cite{sun2019unsupervised, carlucci2019domain, xu2019self}. Xu et al. \cite{xu2019self} suggested using self-supervision pretext tasks~(\textit{e.g.} image rotation, patch location prediction) over a feature extractor. Feng et al.  \cite{feng2019self} proposed to use self-supervision pretext tasks as part of their framework for domain generalization. 
Carlucci et al. \cite{carlucci2019domain} proposed to solve domain adaptation/generalization by introducing a jigsaw puzzle as a self-supervision task. Images are decomposed into 9 patches which are then randomly shuffled and used to form images of the same dimension of the original ones. The Maximal Hamming distance algorithm is used to define a set of patch permutations and assign an index to each of them. The convolutional network is optimized to satisfy two objectives: object recognition on the ordered images and jigsaw classification, namely the permutation index recognition, on the shuffled images. Sun et al. \cite{sun2019unsupervised} further proposed to perform domain adaptation by jointly learning multiple self-supervision tasks. Source and target images share the same convolutional feature encoder, and the extracted features are then fed into different self-supervision task heads: image rotation prediction, patch location prediction, and flip prediction. Since images from different domains normally have many low-level visual differences, \textit{e.g.} brightness, texture, etc.,  self-supervision tasks aiming to predict pixel values of the original images are usually not quite helpful. Because of this, self-supervision tasks that predict high-level structural labels are more favorable for domain adaptation. Kim et al. \cite{kim2020cross} proposed a cross-domain self-supervised learning approach for DA. It captures apparent visual similarities with both in-domain and across-domain self-supervision. Consequently, they could perform DA with only few source labels. Self-supervised learning has also been introduced into point-cloud adaptation~\cite{achituve2020self}, in which region reconstruction is introduced as a new pretext task.


\subsection{Combinations and Others}
\label{ssec:Others}

Some techniques combine several of the above-mentioned methods to jointly explore their advantages. Zhang et al. \cite{zhang2018fully} performed adaptation in both the visual appearance-level and representation-level. Leveraging the unpaired image-to-image translation framework~\cite{zhu2017unpaired}, the method proposed by Murez et al. \cite{murez2018image} requires that the extracted features are able to reconstruct the images in both domains. In addition, they also aligned the extracted features in both domains.

Finding invariant representations alone is clearly not a sufficient condition for the success of domain adaptation. Zhao et al. \cite{zhao2019learning} gave a simple counterexample where invariant representations lead to large joint error on source and target domains. So far, most methods focus on covariate shift 
which occurs on standard datasets but fails in most practical applications. For instance, when transferring knowledge from synthetic to real images \cite{peng2017visda}, the supports of the input distributions are actually disjoint. Similar to covariate shift, label shift is also a long-standing problem in machine learning, but only a few works in domain adaptation have focused on solving it until recently. In this line of work, \cite{zhao2019learning, combes2020domain, lipton2018detecting, tan2019generalized, azizzadenesheli2019regularized} proposed generalization bounds for this scenario and focused on detection and alignment by estimating the density ratio $P(y_s) / P(y_t)$ and doing importance re-sampling on $\mathcal{Y}$ space.

\begin{table*}[!t]
\centering
\caption{Comparison of different single-source DUDA categories. (The more stars the method has, the better it is. )}
\resizebox{0.9\textwidth}{!}{%
\begin{tabular}{c|c|c|c|c|c|c|c}
\toprule
  & Theory guarantee & Efficiency & Task scalability & Data scalability & Data dependency & Optimizability & Performance \\ \hline
  Discrepancy-based methods & $\bigstar\bigstar\bigstar$ & $\bigstar\bigstar\bigstar$ & $\bigstar$ & $\bigstar\bigstar$ & $\bigstar\bigstar\bigstar$ & $\bigstar\bigstar\bigstar$ & $\bigstar\bigstar$  \\ \hline
  Adversarial discriminative methods & $\bigstar\bigstar$ & $\bigstar\bigstar$ & $\bigstar\bigstar\bigstar$ & $\bigstar\bigstar\bigstar$ & $\bigstar$ & $\bigstar$ & $\bigstar\bigstar\bigstar$ \\ \hline
  Adversarial generative methods & $\bigstar$ & $\bigstar$ & $\bigstar\bigstar$ & $\bigstar$ & $\bigstar$ & $\bigstar$ & $\bigstar\bigstar\bigstar$ \\ \hline
  Self-supervision methods & $\bigstar$ & $\bigstar\bigstar$ & $\bigstar\bigstar\bigstar$ & $\bigstar\bigstar\bigstar$ & $\bigstar\bigstar$ & $\bigstar\bigstar\bigstar$ & $\bigstar\bigstar$ \\ \bottomrule
\end{tabular}
}
\label{tab:comparison}
\end{table*}

\begin{table}[!t]
\centering
\caption{Performance Comparison (Classification Accuracy in \%) of Different Methods on Digit Dataset for digit recognition. `Backbone' denotes the base network architecture, `M', `M-M', `U', `S' are different domains (see Section~\ref{sec:Datasets} for details), and `-->' represents the adaptation from one source domain to another target domain. The column `\textbf{C}' indicates which category the method belongs to, where `\textbf{D}', `\textbf{A}', `\textbf{G}', `\textbf{S}', `\textbf{O}' are respectively short for discrepancy-based, adversarial discriminative, adversarial generative, self-supervision-based methods, and others (the same below).}
\resizebox{0.48\textwidth}{!}{%
\begin{tabular}{llccccccc}
\toprule
Backbone                  & Method                                  & Venue  & \textbf{C}  & M-->M-M & M-->U & U-->M & S-->M & M-->S\\ \midrule
AlexNet                   & SWDA~\cite{rozantsev2018beyond}          & TPAMI 2018    & \textbf{D} & -               & 60.7         & 67.3         & -            & -            \\\midrule
\multirow{11}{*}{Custom}  & DANN~\cite{ganin2016domain}              & JMLR 2016   & \textbf{A}   & 76.7            & -            & -            & 73.9         & -            \\
& DRCN~\cite{ghifary2016deep}              & ECCV 2016    & \textbf{S}   & -               & 91.8         & 73.7        & 82.0        & 40.1        \\
& DSN~\cite{bousmalis2016domain}           & NeurIPS 2016  & \textbf{S} & 83.2           & -            & -            & 82.7        & -            \\
& PixelDA~\cite{bousmalis2017unsupervised} & CVPR 2017   & \textbf{G}    & 98.2            & 95.9         & -            & -            & -            \\
& UNIT~\cite{liu2017unsupervised}          & NeurIPS 2017   & \textbf{G}   & -               & 96.0        & 93.6        & 90.5        & -            \\
& CyCADA~\cite{hoffman2018cycada}          & ICML 2018   & \textbf{G}   & -               & 95.6         & 96.5         & 90.4         &              \\
& SEDA~\cite{french2018self}               & ICLR 2018   & \textbf{O}   & -               & 98.2            & 99.6        & 99.3        & 97.0           \\
& MCD~\cite{saito2018maximum}              & CVPR 2018   & \textbf{O}   & -               & 94.2         & 94.1         & 96.2         & -            \\
& SWD~\cite{lee2019sliced}                 & CVPR 2019   & \textbf{D}   & -               & 98.1         & 97.1         & 98.9         & -            \\
& DWT~\cite{roy2019unsupervised}       & CVPR 2019   & \textbf{D}   & -               & 99.1         & 98.8        & 97.8        & 28.9        \\ 
 & RCA~\cite{cicek2019unsupervised}         & ICCV 2019   & \textbf{A}    & 99.5           & -            & -            & 99.3        & 89.2        \\
\midrule
ResNet26                  & SSDA~\cite{sun2019unsupervised}          & arXiv 2019    & \textbf{S}      & 98.9            & 96.5         & 90.2         & 85.8         & 61.3         \\ \midrule
\multirow{2}{*}{ResNet50} & CDAN~\cite{long2018conditional}          & NeurIPS 2018   & \textbf{A}   & -               & 95.6         & 98.0           & 89.2         & -            \\ 
& 3CATN~\cite{li2019cycle}                 & ACM MM 2019     & \textbf{G}   & -               & 96.1  & 98.3   & 92.5   & -            \\ \midrule
\multirow{4}{*}{LeNet}    & ADDA~\cite{tzeng2017adversarial}         & CVPR 2017   & \textbf{A}    & -               & 89.4         & 90.1         & 76.0           & -            \\
& I2I~\cite{murez2018image}                & CVPR 2018   & \textbf{O}   & -               & 98.8         & 97.6         & 90.1         & -            \\
& TPN~\cite{pan2019transferrable}          & CVPR 2019    & \textbf{O}   & -               & 92.1         & 94.1         & 93.0           & -            \\
& HoMM~\cite{chen2020homm}                 & AAAI 2020    & \textbf{D}  & -               & -            & 99.1         & 99.0           & -            \\ \bottomrule
\end{tabular}
}
\label{tab:digits}
\end{table}

Recently, pseudo-labeling has been exploited in a number of DA methods. Pan et al. \cite{pan2019transferrable} and Hu et al. \cite{hu2020panda} assigned pseudo-labels to images in the target domain and then performed domain alignment based on prototypes. These methods are mostly used for image classification. Zou et al. \cite{zou2018unsupervised} proposed to utilize pseudo-labeling in self-training for semantic segmentation, in which pseudo-labels are generated from high-confidence predictions. However, since pseudo-labels are noisy, overconfident label belief can be paced on wrong classes, leading to propagated errors. In order to solve this issue, Zou et al. \cite{zou2019confidence} proposed a confidence regularized self-training framework, in which pseudo-labels are treated as continuous latent variables jointly optimized via alternating optimization. Label regularization and model regularization are proposed as two types of confidence regularizations.

Ensemble methods have also been used for DA. \cite{laine2017temporal, tarvainen2017mean} originally developed ensemble methods for semi-supervised learning. Laine et al. \cite{laine2017temporal} performed ensembling by averaging over past predictions for each example, while Tarvainen et al. \cite{tarvainen2017mean} performed ensemble by leveraging past network weights. These ensemble approaches require high randomness in either inputs or network models, which can be provided by data augmentation, varying augmentation parameters, and utilizing dropout. French et al. \cite{french2018self} extended these methods for unsupervised domain adaptation. Images are first processed with stochastic data augmentation and then fed into both networks. The student network is trained with gradient descent while the teacher network updates its weights using an exponential moving average of the student network's weights. Stochastic weight averaging   further improves the adaptation results as shown in~\cite{athiwaratkun2019there}. Recently, Cai et al. \cite{cai2019exploring} proposed Mean Teacher with Object Relations~(MTOR) for object detection which integrates object relations into the consistency cost between teacher and student modules.

Other techniques are quite different from what we have mentioned above. Zhang et al. \cite{zhang2017curriculum} proposed a curriculum-style learning approach to minimize the domain gap in semantic segmentation. This method solves easy tasks first in order to infer some necessary properties about the target domain and then the network predictions in the target domain are enforced to follow those inferred properties during the training process. Wang et al. \cite{wang2018visual} proposed a Manifold Embedded Distribution Alignment approach which learns a domain-invariant classifier in Grassmann manifold with structural risk minimization, while performing dynamic distribution alignment to quantitatively account for the relative importance of marginal and conditional \mbox{distributions}. Saito et al. \cite{saito2018maximum} introduced a new approach that attempts to align the distributions of the source and target domains by utilizing the task-specific decision boundaries. They proposed to maximize the discrepancy between two classifiers' outputs to detect target samples that are far from the support of the source. Khodabandeh et al. \cite{khodabandeh2019robust} recently proposed to address DA from the perspective of robust learning and showed that the problem may be ormulated as training with noisy labels. 
Chen et al. \cite{chen2018road} proposed a target guided distillation approach to learn the real image style, which is achieved by training the segmentation model to imitate a pre-trained real style model using real images. They further took advantage of the intrinsic spatial structure presented in urban scene images, and proposed a spatial-aware adaptation scheme to effectively align the distribution of two domains.

\begin{table}[!t]
\centering
\caption{Performance Comparison (Classification Accuracy in \%) of Different Methods on Office-31 Datasets for object classification. `A', `D', `W' are different domains in the Office-31 dataset, and `Avg' is the average performance of different adaptation settings (the same below).}
\resizebox{0.5\textwidth}{!}{%
\begin{tabular}{llccccccccc}
\toprule
Backbone                    & Method                                 & Venue  & \textbf{C} & A-->W & D-->W & W-->D & A-->D & D-->A & W-->A & Avg \\ \midrule
\multirow{10}{*}{AlexNet}                       & DANN~\cite{ganin2016domain}             & JMLR 2016  & \textbf{A} & 73.0    & 96.4  & 99.2  & -     & -     & -     & -       \\ 
& SWDA~\cite{rozantsev2018beyond}         & TPAMI 2018  & \textbf{D} &  76.0    & 96.7  & 99.6  & -     & -     & -     & -       \\
& CORAL~\cite{sun2017correlation}         & \begin{tabular}[c]{@{}c@{}}DACVA 2017 \end{tabular}  & \textbf{D} &  66.4  & 95.7  & 99.2  & 66.8  & 52.8  & 51.5  & 72.1    \\
& DAN~\cite{long2015learning}             & JMLR 2015  & \textbf{D} & 68.5  & 96.0    & 99.0    & 67.0    & 54.0    & 53.1  & 72.9    \\
& DUCDA~\cite{zhuo2017deep}               & ACM MM 2017  & \textbf{D} &  68.3  & 96.2  & 99.7  & 68.3  & 53.6  & 51.6  & 73.0      \\
& DRCN~\cite{ghifary2016deep}             & ECCV 2016  & \textbf{S} & 68.7  & 96.4  & 99.0    & 66.8  & 56.0    & 54.9  & 73.6    \\
& JMMD~\cite{long2017deep}                & ICML 2017  & \textbf{D} &  75.2  & 96.6  & 99.6  & 72.8  & 57.5  & 56.3  & 76.3    \\
& AutoDIAL~\cite{cariucci2017autodial}    & ICCV 2017  & \textbf{D} &  75.5  & 96.6  & 99.5  & 73.6  & 58.1  & 59.4  & 77.1    \\
& CDAN~\cite{long2018conditional}         & NeurIPS 2018  & \textbf{A} & 78.3  & 97.2  & 100.0   & 76.3  & 57.3  & 57.3  & 77.7    \\
& DM-ADA~\cite{xu2019adversarial}         & AAAI 2020  & \textbf{A} &  83.9  & 99.8  & 99.9  & 77.5  & 64.6  & 64.0    & 81.6    \\
 \midrule
\multirow{11}{*}{ResNet-50} 
& SSDA~\cite{xu2019self}                  & Access 2019  & \textbf{S} &  88.6  & 98.0    & 100.0   & 85.7  & 68.0    & 65.5  & 84.3    \\
& JMMD~\cite{long2017deep}                & ICML 2017  & \textbf{D} &  86.0    & 96.7  & 99.7  & 85.1  & 69.2  & 70.7  & 84.6    \\
& HoMM~\cite{chen2020homm}                & AAAI 2020  & \textbf{D} &  90.8  & 99.3  & 100.0   & 87.9  & 69.3  & 69.5  & 86.1    \\
& GTA~\cite{sankaranarayanan2018generate} & CVPR 2018  & \textbf{G} &  89.5  & 97.9  & 99.8  & 87.7  & 72.8  & 71.4  & 86.5    \\
& CRST~\cite{zou2019confidence}           & ICCV 2019  & \textbf{O} &  89.4  & 98.9  & 100.0   & 88.7  & 72.6  & 70.9  & 86.8   \\
& DAAA~\cite{kang2018deep}                & ECCV 2018 & \textbf{G}  &  86.8  & 99.3  & 100.0   & 88.8  & 74.3  & 73.9  & 87.2    \\
& CDAN~\cite{long2018conditional}         & NeurIPS 2018  & \textbf{A} &  94.1  & 98.6  & 100.0   & 92.9  & 71.0    & 69.3  & 87.7    \\
& TAT~\cite{liu2019transferable}          & ICML 2019  & \textbf{A} &  92.5  & 99.3  & 100.0   & 93.2  & 73.1  & 72.1  & 88.4    \\
& AL2DA~\cite{chen2020adversarial}        & AAAI 2020  & \textbf{A} &  95.6  & 97.7  & 100.0   & 94    & 72.2  & 72.5  & 88.7    \\
& 3CATN~\cite{li2019cycle}                & ACM MM 2019  & \textbf{G} & 95.3  & 99.3  & 100.0 &  94.1  & 73.1  & 71.5    & 88.9    \\
& PANDA~\cite{hu2020panda}                & arXiv 2020  & \textbf{A} &  94.9  & 97.8  & 99.8  & 94.2  & 73.9  & 72.8  & 88.9    \\ 
& CAN~\cite{kang2019contrastive}          & CVPR 2019  & \textbf{D} &  94.5  & 99.1  & 99.8  & 95    & 78.0    & 77.0    & 90.6    \\
                            \midrule
VGG16                       & CMD~\cite{zellinger2017central}         & ICLR 2017  & \textbf{D} &  77.0    & 96.3  & 99.2  & 79.6  & 63.8  & 63.3  & 79.9    \\ \midrule
\multirow{2}{*}{Inception}  & ABN-DA~\cite{li2018adaptive}            & PR 2018  & \textbf{D} &  75.4  & 96.2  & 99.6  & 72.7  & 59.0    & 60.5  & 77.2\\ 
& AutoDIAL~\cite{cariucci2017autodial}    & ICCV 2017  & \textbf{D} &  84.2  & 97.9  & 99.9  & 82.3  & 64.6  & 64.2  & 82.2    \\
\midrule
ResNet-34                   & I2I~\cite{murez2018image}               & CVPR 2018  & \textbf{O} &  75.3  & 96.5  & 99.6  & 71.1  & 50.1  & 52.1  & 74.1    \\    \bottomrule
\end{tabular}%
}
\label{tab:office-31}
\end{table}

\begin{table*}[]
\centering
\caption{Performance Comparison (Classification Accuracy in \%) of Different Methods on Office-Home Dataset for object classification.  `Ar', `Cl', `Pr', `Rw' are different domains in the Office-Home dataset.}
\resizebox{0.95\textwidth}{!}{%
\begin{tabular}{llccccccccccccccc}
\toprule
BackBone & Method & Venue  & \textbf{C} & Ar-->Cl & Ar-->Pr & Ar-->Rw & Cl-->Ar & Cl-->Pr & Cl-->Rw & Pr-->Ar & Pr-->Cl & Pr-->Rw & Rw-->Ar & Rw-->Cl & Rw-->Pr & Avg \\ \midrule
AlexNet & CDAN~\cite{long2018conditional} & NeurIPS 2018  & \textbf{A} & 38.1    & 50.3    & 60.3    & 39.7    & 56.4    & 57.8    & 35.5    & 43.1    & 63.2    & 48.4    & 48.5    & 71.1    & 51.0 \\ \midrule
\multirow{6}{*}{ResNet-50}  & HoMM~\cite{chen2020homm} & AAAI 2020  & \textbf{D} & -       & 64.7    & 71.8    & -       & -       & 66.1    & -       & -       & 74.5    & -       & -       & 81.2    & -       \\
 & DWT-MEC~\cite{roy2019unsupervised} & CVPR 2019  & \textbf{D} & 50.3    & 72.1    & 77.0      & 59.6    & 69.3    & 70.2    & 58.3    & 48.1    & 77.3    & 69.3    & 53.6    & 82.0      & 65.6 \\
& CDAN~\cite{long2018conditional} & NeurIPS 2018  & \textbf{A} & 50.7    & 70.6    & 76.0      & 57.6    & 70.0      & 70.0      & 57.4    & 50.9    & 77.3    & 70.9    & 56.7    & 81.6    & 65.8    \\
 & TAT~\cite{liu2019transferable} & ICML 2019  & \textbf{A} & 51.6    & 69.5    & 75.4    & 59.4    & 69.5    & 68.6    & 59.5    & 50.5    & 76.8    & 70.9    & 56.6    & 81.6    & 65.8    \\
 & AL2DA~\cite{chen2020adversarial} & AAAI 2020  & \textbf{A} & 53.7    & 70.1    & 76.4    & 60.2    & 72.6    & 71.5    & 56.8    & 51.9    & 77.1    & 70.2    & 56.3    & 82.1    & 66.6    \\
 & PANDA~\cite{hu2020panda} & arXiv 2020   & \textbf{A}    & 52.4    & 73.4    & 79.0      & 64.2    & 74.2    & 73.2    & 63.0      & 53.0      & 79.5    & 73.4    & 56.7    & 83.5    & 68.8    \\
\bottomrule
\end{tabular}%
}
\label{tab:office-home}
\end{table*}

\subsection{Qualitative Comparison (Table~\ref{tab:comparison})}
\label{ssec:qualitative}

To thoroughly review the various single-source DUDA methods, we use the following qualitative criteria: 
1) \textit{Theory guarantee}: if the target risk has upper bound; and if the upper bound can be minimized by the algorithm. 2) \textit{Efficiency}: the computation cost of the training and inference of the algorithm. 3) \textit{Task scalability}: if the algorithm is applicable to complex tasks, such as semantic segmentation and object detection. 4) \textit{Data scalability}: if the algorithm is applicable to large and complex datasets with rather diversified images. 5) \textit{Data dependency}: if the algorithm can be well trained with small datasets. 6) \textit{Optimizability}: if the algorithm is easy to train and requires less hyper-parameter tuning. 7) \textit{Performance}: how well the algorithm performs. 

\begin{table}[!t]
\centering
\caption{Performance Comparison (Classification Accuracy in \%) of Different Methods on VisDA-2017 Dataset for object classification. The simulation domain (sim) and the real-world domain (real) are respectively used as source and target.}
\resizebox{0.4\textwidth}{!}{%
\begin{tabular}{llccc}
\toprule
BackBone                    & Method                                                            & Venue  & \textbf{C} & Sim->Real \\ \midrule
\multirow{5}{*}{ResNet-50}  & DAN~\cite{long2015learning}                                        & ICML 2015  & \textbf{D} & 63.7       \\
& GTA~\cite{sankaranarayanan2018generate}                            & CVPR 2018  & \textbf{G} & 69.5       \\
& CDAN~\cite{long2018conditional}                                    & NeurIPS 2018  & \textbf{A} & 70.0         \\
& TAT~\cite{liu2019transferable}                                     & ICML 2019  & \textbf{A} & 71.9       \\
& 3CATN~\cite{li2019cycle}                                           & ACM MM 2019    & \textbf{G} & 73.2       \\ \midrule
\multirow{7}{*}{ResNet-101} & DAN~\cite{long2015learning}                                                               & ICML 2015  & \textbf{D} & 62.8       \\
& DM-ADA~\cite{xu2019adversarial}                                    & AAAI 2020   & \textbf{A}  & 75.6       \\
& SWD~\cite{lee2019sliced}                                           & CVPR 2019  & \textbf{D} & 76.4       \\ 
& CRST~\cite{zou2019confidence}                                      & ICCV2019  & \textbf{O}  & 78.1       \\ 
& PANDA~\cite{hu2020panda}                                           & arXiv 2020  & \textbf{A}    & 78.3       \\ 
& self-ensembling~\cite{french2018self} & ICLR 2018  & \textbf{O} & 82.8       \\
& CAN~\cite{kang2019contrastive}                                     & CVPR 2019  & \textbf{D} & 87.2       \\
\bottomrule
\end{tabular}%
}
\label{tab:visda-2017}
\end{table}

Discrepancy-based methods usually define a distance measurement between the source and target distributions. Based on this definition, an upper bound of the target risk can be derived and  domain adaptation algorithms can be designed to minimize this upper bound. Compared with other DUDA categories, many of the existing discrepancy-based methods have better theoretical guarantees. Since most discrepancy-based methods do not add significantly large blocks onto the backbone network, the whole network architectures are usually not very complicated. On the one hand, the computation efficiency of the discrepancy-based methods is usually higher than other categories and the training of the network does not highly rely on large datasets. On the other hand, these methods are not as applicable to large and complex datasets with more diversified images as other categories. In terms of optimizability, since the networks are not very complicated, they are easier to train and require less hyperparameter tuning. Most of the discrepancy-based methods learn image-level representations, instead of pixel-level ones, thus they are not as applicable to complex tasks, such as semantic segmentation, as other categories. It is difficult for most discrepancy-based methods to achieve satisfying performance on complex datasets and tasks.

Adversarial discriminative approaches are the most widely used methods to solve DA problems and achieve remarkable results. Several theoretical studies on these methods focus on the investigation of generalization bound and risk analysis. These methods have competitive computational efficiency and task scalability. In terms of data scalability, they work well across different kinds of datasets. Due to the reliance on the convergence of a min-max game between the discriminator and the feature extractor, they do not always work well on small datasets and are also relatively difficult to optimize.

There is usually no good theoretical support behind adversarial generative approaches since they mainly leverage GAN or other kinds of generative models to reduce the visual gap between source and target domains. However, they usually perform well on many complex tasks with high dimensional solution space, such as semantic segmentation and object detection. It is also because of their reliance on the generative models that they usually require the source and target domains to have homogeneous visual patterns and cannot easily scale to more complex datasets. Since they rely on generative models to build pattern transformation between source and target domains, they require large-scale datasets to robustly train the generative model. Correspondingly, these approaches also require  more computing resources and a more complicated optimization process.

\begin{table}[!t]
\centering
\caption{Performance comparison (in \%) of different methods from Cityscapes to KITTI for object detection. The 4th to the 8th columns indicate the Average Precision (AP) for the 5 different classes, and the last column is the mean Average Precision (mAP).}
\resizebox{0.48\textwidth}{!}{%
\begin{tabular}{llccccccccc}
\toprule
BackBone                & Method                    & Venue    & \textbf{C}      & \rot{Person} & \rot{Rider} & \rot{Car}  & \rot{Truck} & \rot{Train} & mAP\\ \midrule
\multirow{3}{*}{VGG-16} & DAF~\cite{chen2018domain}  & CVPR 2018    & \textbf{A}       & 40.9   & 16.1  & 70.3 & 23.6  & 21.2  & 34.4  \\
& MDA~\cite{xie2019multi}    & ICCVW 2019  & \textbf{A} & 53.3   & 24.5  & 72.2 & 28.7  & 25.3  & 40.7  \\
& CFFA~\cite{zheng2020cross} & arXiv 2020  & \textbf{A}           & 50.4   & 29.7  & 73.6 & 29.7  & 21.6  & 41.0 \\ \bottomrule
\end{tabular}%
}
\label{tab:CSKITTI}
\end{table}

\begin{table*}[!t]
\centering
\caption{
Performance comparison (in \%) of different methods from Cityscapes to FoggyCityscapes for object detection. The 4th to the 11st columns indicate the Average Precision (AP) for the 8 different classes.
}
\resizebox{0.75\textwidth}{!}{%
\begin{tabular}{llcccccccccccc}
\toprule
BackBone                & Method                           & Venue     & \textbf{C}     & Bus  & Bicycle & Car  & Motor & Person & Rider & Train & Truck & mAP  \\ \midrule
ResNet-50               & MTOR~\cite{cai2019exploring}      & CVPR 2019     & \textbf{O}      & 38.6 & 35.6    & 44.0   & 28.3  & 30.6   & 41.4  & 40.6  & 21.9  & 35.1 \\ \midrule
Inception-v2            & RLDA~\cite{khodabandeh2019robust} & ICCV 2019    & \textbf{O}       & 45.3 & 36.0      & 49.2 & 26.9  & 35.1   & 42.2  & 27.0    & 30.0    & 36.5  \\ \midrule
\multirow{8}{*}{VGG-16} & DAF~\cite{chen2018domain}         & CVPR 2018    & \textbf{A}       & 35.3 & 27.1    & 40.5 & 20.0    & 25.0     & 31.0    & 20.2  & 22.1  & 27.6 \\
& SCDA~\cite{zhu2019adapting}       & CVPR 2019     & \textbf{A}      & 39.0   & 33.6    & 48.5 & 28.0    & 33.5   & 38.0    & 23.3  & 26.5  & 33.8 \\
& MAF~\cite{he2019multi}            & ICCV 2019     & \textbf{A}      & 39.9 & 33.9    & 43.9 & 29.2  & 28.2   & 39.5  & 33.3  & 23.8  & 34.0   \\
& SWDA~\cite{saito2019strong}       & CVPR 2019    & \textbf{A}       & 36.2 & 35.3    & 43.5 & 30.0    & 29.9   & 42.3  & 32.6  & 24.5  & 34.3  \\
& DD-MRL~\cite{kim2019diversify}    & CVPR 2019     & \textbf{G}      & 38.4 & 32.2    & 44.3 & 28.4  & 30.8   & 40.5  & 34.5  & 27.2  & 34.6  \\
& MDA~\cite{xie2019multi}           & ICCVW 2019  & \textbf{A} & 41.8 & 36.5    & 44.8 & 30.5  & 33.2   & 44.2  & 28.7  & 28.2  & 36.0   \\
& PDA~\cite{hsu2020progressive}     & WACV 2020  & \textbf{G} & 44.1 & 35.9    & 54.4 & 29.1  & 36.0     & 45.5  & 25.8  & 24.3  & 36.9  \\
& CFFA~\cite{zheng2020cross}        & arXiv 2020     & \textbf{A}        & 43.2 & 37.4    & 52.1 & 34.7  & 34.0     & 46.9  & 29.9  & 30.8  & 38.6 \\ \bottomrule
\end{tabular}%
}
\label{tab:CSFCS}
\end{table*}

Despite the apparent difference, both discrepancy-based methods and adversarial methods can be understood as approaches that attempt to align the marginal feature distributions of both domains. While both methods are intuitive and have seen empirical success in several cases, fundamental limitation exists for both lines of work.

In a recent paper~\citep{zhao2019learning}, the authors proved an information-theoretic lower bound on the joint error of methods based on learning domain-invariant representations, showing that when the label distributions of the two domains differ, any algorithm has to achieve a large error on at least one of the two domains. Since only source error could be minimized due to the availability of labeled samples, this implies an increasing error on the target domain. Furthermore, the better the distribution alignment, the worse the joint error. In a concurrent work,~\citet{johansson2019support} also identified the insufficiency of learning domain-invariant representation for successful adaptation. They further analyzed the information loss of non-invertible transformations, and proposed a generalization upper bound that directly takes it into account.

While most of the work we discussed so far focuses on learning domain-invariant representations, methods based on estimating the importance ratio of density functions between target and source domains are also abundant in the literature~\citep{huang2007correcting,sugiyama2007covariate,gretton2009covariate,yu2012analysis,adel2017unsupervised}. Most of these approaches exhibit provable generalization guarantees under the covariate shift assumption. An interesting avenue for future research is combining  the distribution alignment method using deep networks for feature learning with importance ratio reweighting. Note that, different from traditional methods where the importance ratio is estimated between the data density functions, recent work has explored the alternative direction where the importance ratio between the marginal label distributions of the two domains is estimated instead~\citep{lipton2018detecting,azizzadenesheli2019regularized}. 
The fundamental limitation of domain-invariant representations is the potential discrepancy between the marginal label distributions. To overcome such lower bound, one could use the importance ratio between label distributions to compensate for such label discrepancy, as explored in several recent work~\citep{li2019target,combes2020domain}.

Compared with other methods, self-supervision-based methods do not have a strong theoretical guarantee since these methods are mostly based on the intuition that by forcing the CNN encoder to perform the desired task on the source domain and the pretext tasks on the target domain, the CNN encoder could extract domain-invariant features for both. In terms of computation cost, self-supervision-based methods perform the self-supervision tasks with additional heads, which are normally light-weight CNNs. They normally have more computation cost than discrepancy-based methods, while having less computation cost than adversarial generative methods. Self-supervision-based methods do not have assumptions on the downstream task, and are applicable to complex tasks. In terms of data scalability, self-supervision-based methods are robust and applicable to
complex datasets. The self-supervision tasks are normally simple tasks which are easy to train along with the downstream task network. Finally, self-supervision-based methods usually have better performance than discrepancy-based methods, but are less performant than adversarial discriminative and generative methods.

\subsection{Quantitative Comparison (Table VIII to Table XV)}
\label{ssec:quantitative}

In this section, we quantitatively compare different categories of single-source DUDA methods in three visual tasks, \textit{i.e.} image classification, object detection, and semantic segmentation. First, we introduce detailed experimental settings, including datasets with their properties, and evaluation metrics. Second, we analyze the results.

\subsubsection{Image Classification}
We compare the classification performance of different methods on 4 different datasets, Digit, Office-31, Office-Home, and VisDA-2017. The first three datasets contain several domains of images. A DA method is evaluated by performing adaptation from each domain to every other domain in the dataset, and averaging all adaptation performances.  Classification accuracy is used as the evaluation metric. 

Digit and Office-31 are relatively basic datasets for DA. Because images in these datasets are mostly centered objects with simple backgrounds, many methods could achieve high adaptation accuracy, making it hard to compare them. However, these datasets are still widely used since they are convenient for testing new ideas. Office-Home contains more domains~(4) and the 12 source-to-target adaptation settings provide more diverse tests to mitigate the possibility of over-fitting. VisDA-2017 is a challenging large-scale dataset with one simulation domain and one real-world domain.

\begin{table*}[!t]
\centering
\caption{Performance Comparison (in \%) of Different Methods from GTA to Cityscapes for semantic segmentation. The 4th to the 22nd columns indicate the class-wise intersection-over-union (cwIoU) for the 19 different classes, and the last column is the mean intersection-over-union (mIoU).}
\resizebox{0.95\textwidth}{!}{%
\begin{tabular}{llcccccccccccccccccccccc}
\toprule
BackBone                    & Method                             & Venue  & \textbf{C}   & \rot{road} & \rot{sidewalk} & \rot{building} & \rot{wall} & \rot{fence} & \rot{pole} & \rot{traffic light} & \rot{traffic sign} & \rot{vegettion} & \rot{terrain} & \rot{sky}   & \rot{person} & \rot{rider} & \rot{car}  & \rot{truck} & \rot{bus}  & \rot{train} & \rot{motorbike} & \rot{bicycle} & \rot{mIoU}  \\ \midrule
\multirow{8}{*}{VGG16}      & FCN-Wild~\cite{hoffman2016fcns}     & arXiv 2016    & \textbf{A}    & 70.4 & 32.4     & 62.1     & 14.9 & 5.4   & 10.9 & 14.2          & 2.7          & 79.2       & 21.3    & 64.6 & 44.1   & 4.2   & 70.4 & 8     & 7.3  & 0.0     & 3.5      & 0.0       & 27.1  \\ 
& MCD~\cite{saito2018maximum}         & CVPR 2018   & \textbf{O}  & 86.4 & 8.5      & 76.1     & 18.6 & 9.7   & 14.9 & 7.8           & 0.6          & 82.8       & 32.7    & 71.4 & 25.2   & 1.1   & 76.3 & 16.1  & 17.1 & 1.4   & 0.2      & 0.0       & 28.8  \\
& CDA~\cite{zhang2017curriculum}    & ICCV 2017  & \textbf{O}   & 74.9 & 22.0       & 71.7     & 6.0    & 11.9  & 8.4  & 16.3          & 11.1         & 75.7       & 13.3    & 66.5 & 38.0     & 9.3   & 55.2 & 18.8  & 18.9 & 0.0     & 16.8     & 14.6    & 28.9  \\
& AdaptSegNet~\cite{tsai2018learning} & CVPR 2018  & \textbf{A}   & 87.3 & 29.8     & 78.6     & 21.1 & 18,2  & 22.5 & 21.5          & 11.0           & 79.7       & 29.6    & 71.3 & 46.8   & 6.5   & 80.1 & 23.0    & 26.9 & 0.0     & 10.6     & 0.3     & 35.0    \\
& CyCADA~\cite{hoffman2018cycada}     & ICML 2018  & \textbf{G}   & 85.2 & 37.2     & 76.5     & 21.8 & 15.0    & 23.8 & 22.9          & 21.5         & 80.5       & 31.3    & 60.7 & 50.5   & 9.0     & 76.9 & 17.1  & 28.2 & 4.5   & 9.8      & 0.0       & 35.4  \\
& ROAD~\cite{chen2018road}            & CVPR 2018  & \textbf{O}   & 85.4 & 31.2     & 78.6     & 27.9 & 22.2  & 21.9 & 23.7          & 11.4         & 80.7       & 29.3    & 68.9 & 48.5   & 14.1  & 78.0   & 19.1  & 23.8 & 9.4   & 8.3      & 0.0       & 35.9  \\
& DCAN~\cite{wu2018dcan}                                & ECCV 2018   & \textbf{G}      & 82.3 & 26.7     & 77.4     & 23.7 & 20.5  & 20.4 & 30.3          & 15.9         & 80.9       & 25.4    & 69.5 & 52.6   & 11.1  & 79.6 &   24.9  & 21.2 &  1.3   & 17.0     & 6.7   & 36.2 \\
& CGAN~\cite{hong2018conditional}                               & CVPR 2018  & \textbf{A}   & 89.2 & 49.0       & 70.7     & 13.5 & 10.9  & 38.5 & 29.4          & 33.7         & 77.9       & 37.6    & 65.8 & 75.1   & 32.4  & 77.8 & 39.2  & 45.2 & 0.0     & 25.5     & 35.4    &  44.5     \\
\midrule
\multirow{8}{*}{ResNet-101} & DCAN~\cite{wu2018dcan}              & ECCV 2018   & \textbf{G}      & 88.5 & 37.4     & 79.3     & 24.8 & 16.5  & 21.3 & 26.3          & 17.4         & 80.8       & 30.9    & 77.6 & 50.2   & 19.2  & 77.7 & 21.6  & 27.1 & 2.7   & 14.3     & 18.1    &   38.5    \\
& ROAD~\cite{chen2018road}            & CVPR 2018  & \textbf{O}   & 76.3 & 36.1     & 69.6     & 28.6 & 22.4  & 28.6 & 29.3          & 14.8         & 82.3       & 35.3    & 72.9 & 54.4   & 17.8  & 78.9 & 27.7  & 30.3 & 4.0     & 24.9     & 12.6    & 39.4  \\
& UDA-SS~\cite{sun2019unsupervised}    & arXiv 2019  & \textbf{S}       & 86.6 & 37.8     & 80.8     & 29.7 & 16.4  & 28.9 & 30.9          & 22.2         & 83.8       & 37.1    & 76.9 & 60.1   & 7.8   & 84.1 & 30.8  & 32.1 & 1.2   & 23.2     & 13.3    & 41.2  \\
& SSDA~\cite{xu2019self}              & Access 2019  & \textbf{S} & 87.6 & 25.7     & 77.5     & 19.8 & 16.8  & 29.0   & 32.1          & 20.5         & 79.9       & 32.9    & 75.3 & 58.2   & 26.0    & 79.0   & 23.3  & 31.6 & 2.1   & 26.9     & 37.7    & 41.2  \\
& AdaptSegNet~\cite{tsai2018learning} & CVPR 2018   & \textbf{A}  & 86.5 & 36.0       & 79.9     & 23.4 & 23.3  & 23.9 & 35.2          & 14.8         & 83.4       & 33.3    & 75.6 & 58.5   & 27.6  & 73.7 & 32.5  & 35.4 & 3.9   & 30.1     & 28.1    & 42.4       \\
& SWD~\cite{lee2019sliced}            & CVPR 2019   & \textbf{D}  & 92.0   & 46.4     & 82.4     & 24.8 & 24.0    & 35.1 & 33.4          & 34.2         & 83.6       & 30.4    & 80.9 & 56.9   & 21.9  & 82.0   & 24.4  & 28.7 & 6.1   & 25.0       & 33.6    & 44.5  \\
& PANDA~\cite{hu2020panda}            & arXiv 2020    & \textbf{A}     & 92.4 & 51.3     & 82.9     & 31.8 & 24.9  & 32.6 & 35.8          & 20.4         & 84.5       & 38.7    & 79.8 & 60.0     & 25.8  & 85.1 & 33.7  & 44.1 & 9.0     & 27.5     & 22.6    & 46.5  \\ 
& FCAN~\cite{zhang2018fully}          & CVPR 2018  & \textbf{O}    & -    & -        & -        & -    & -     & -    & -             & -            & -          & -       & -    & -      & -     & -    & -     & -    & -     & -        & -       & 47.8 \\ \midrule
DRN105                      & MCD~\cite{saito2018maximum}         & CVPR 2018  & \textbf{O}   &90.3 &31.0 &78.5 &19.7 &17.3 &28.6 &30.9 &16.1 &83.7 &30.0 &69.1 &58.5 &19.6 &81.5 &23.8 &30.0 &5.7 &25.7 &14.3 &39.7  \\\midrule
ResNet-34                   & I2I~\cite{murez2018image}           & CVPR 2018   & \textbf{O}   & 85.3 & 38.0       & 71.3     & 18.6 & 16.0    & 18.7 & 12.0            & 4.5          & 72.0         & 43.4    & 63.7 & 43.1   & 3.3   & 76.7 & 14.4  & 12.8 & 0.3   & 9.8      & 0.6     & 31.8  \\\midrule
\multirow{2}{*}{ResNet-38}  & CBST~\cite{zou2018unsupervised}     & ECCV 2018   & \textbf{O}   & 88.0   & 56.2     & 77.0       & 27.4 & 22.4  & 40.7 & 47.3          & 40.9         & 82.4       & 21.6    & 60.3 & 50.2   & 20.4  & 83.8 & 35.0    & 51.0   & 15.2  & 20.6     & 37.0      & 46.2  \\
& CRST~\cite{zou2019confidence}       & ICCV 2019  & \textbf{O}    & 91.7 & 45.1     & 80.9     & 29.0   & 23.4  & 43.8 & 47.1          & 40.9         & 84.0         & 20.0      & 60.6 & 64.0     & 31.9  & 85.8 & 39.5  & 48.7 & 25.0    & 38.0       & 47.0      & 49.8  \\\midrule
DRN26                       & CyCADA~\cite{hoffman2018cycada}     & ICML 2018   & \textbf{G}  & 79.1 & 33.1     & 77.9     & 23.4 & 17.3  & 32.1 & 33.3          & 31.8         & 81.5       & 26.7    & 69.0   & 62.8   & 14.7  & 74.5 & 20.9  & 25.6 & 6.9   & 18.8     & 20.4    & 39.5  \\\midrule
PSPNet                      & DCAN~\cite{wu2018dcan}              & ECCV 2018    & \textbf{G}     & 85.0   & 30.8     & 81.3     & 25.8 & 21.2  & 22.2 & 25.4          & 26.6         & 83.4       & 36.7    & 76.2 & 58.9   & 24.9  & 80.7 & 29.5  & 42.9 & 2.5   & 26.9     & 11.6    & 41.7 \\ \bottomrule
\end{tabular}%
}
\label{tab:GTACS}
\end{table*}

\begin{table*}[!t]
\centering
\caption{
Performance Comparison (in \%) of Different Methods from SYNTHIA to Cityscapes for semantic segmentation. The 4th to the 19th columns indicate the cwIoU for the 16 different classes, and the last two columns are the mIoU over all the 16 classes and over 13 classes excluding the 3 classes marked with *.
}
\resizebox{0.95\textwidth}{!}{%
\begin{tabular}{llcccccccccccccccccccc}
\toprule
BackBone                    & Method                             & Venue  & \textbf{C} & \rot{road} & \rot{sidewalk} & \rot{building} & \rot{wall*} & \rot{fence*} & \rot{pole*} & \rot{traffic light} & \rot{traffic sign} & \rot{vegetation} & \rot{sky}   & \rot{person} & \rot{rider} & \rot{car} & \rot{bus} & \rot{motorbike} & \rot{bicycle} & \rot{mIoU} & \rot{mIoU*}\\ \midrule
\multirow{6}{*}{VGG16}      & AdaptSegNet~\cite{tsai2018learning} & CVPR 2018  & \textbf{A} & 78.9 & 29.2     & 75.5    &  -    &   -    &    -  & 0.1           & 4.8          & 72.6       & 76.7 & 43.4   & 8.8   & 71.1 & 16.0   & 3.6      & 8.4     & - & 37.6 \\
& FCN-Wild~\cite{hoffman2016fcns}  & arXiv 2016   & \textbf{A}   & 11.5 & 19.6     & 30.8    & 4.4  & 0.0     & 20.3 & 0.1           & 11.7         & 42.3       & 68.7 & 51.2   & 3.8   & 54.0   & 3.2  & 0.2      & 0.6     & 17.0   &22.9 \\
& CDA~\cite{zhang2017curriculum}    & ICCV 2017  & \textbf{O} & 65.2 & 26.1     & 74.9    & 0.1  & 0.5   & 10.7 & 3.7           & 3.0            & 76.1       & 70.6 & 47.1   & 8.2   & 43.2 & 20.7 & 0.7      & 13.1    &   29.0   & 34.8\\ 
& DCAN~\cite{wu2018dcan}                                & ECCV 2018    & \textbf{G}     & 79.9 & 30.4     & 70.8     & 1.6 & 0.6  & 22.3 & 6.7  & 23.0     & 76.9       & 73.9    &  41.9   & 16.7  & 61.7 &   11.5 & 10.3     & 38.6    & 35.4 &41.8 \\
& ROAD~\cite{chen2018road}            & CVPR 2018  & \textbf{O} & 77.7 & 30.0       & 77.5    & 9.6  & 0.3   & 25.8 & 10.3          & 15.6         & 77.6       & 79.8 & 44.5   & 16.6  & 67.8 & 14.5 & 7.0        & 23.8    & 36.2  & 41.7 \\
& CGAN~\cite{hong2018conditional}                               & CVPR 2018  & \textbf{A} & 85.0   & 25.8     & 73.5    & 3.4  & 3.0     & 31.5 & 19.5          & 21.3         & 67.4       & 69.4 & 68.5   & 25.0    & 76.5 & 41.6 & 17.9     & 29.5    &   41.2 & 47.8 \\
\midrule
VGG16-Dilated FCN           & NMD~\cite{chen2017no}               & ICCV 2017  & \textbf{A} & 62.7 & 25.6     & 78.3    &   -   &   -    &   -   & 1.2           & 5.4          & 81.3       & 81.0   & 37.4   & 6.4   & 63.5 & 16.1 & 1.2      & 4.6     & - & 35.7 \\ \midrule
ResNet-38                   & CBST~\cite{zou2018unsupervised}     & ECCV 2018   & \textbf{O} & 53.6 & 23.7     & 75.0      & 12.5 & 0.3   & 36.4 & 23.5          & 26.3         & 84.8       & 74.7 & 67.2   & 17.5  & 84.5 & 28.4 & 15.2     & 55.8    & 42.5 &48.5 \\ \midrule
\multirow{5}{*}{ResNet-101} 
& AdaptSegNet~\cite{tsai2018learning} & CVPR 2018  & \textbf{A} & 84.3 & 42.7     & 77.5    &   -   &   -    &   -   & 4.7           & 7.0            & 77.9       & 82.5 & 54.3   & 21.0    & 72.3 & 32.2 & 18.9     & 32.3    & - &46.7\\
& SWD~\cite{lee2019sliced}            & CVPR 2019  & \textbf{D} & 82.4 & 33.2     & 82.5    & -    & -     & -    & 22.6          & 19.7         & 83.7       & 78.8 & 44.0     & 17.9  & 75.4 & 30.2 & 14.4     & 39.9  & -  & 48.1  \\
& PANDA~\cite{hu2020panda}            & arXiv 2020  & \textbf{A}    & 88.1 & 44.2     & 81.1    & -    & -     & -    & 10.0            & 11.1         & 80.3       & 84.3 & 42.8   & 21.6  & 82.5 & 34.6 & 16.9     & 38.7  & -  & 48.9 \\
& DCAN~\cite{wu2018dcan}                                & ECCV 2018    & \textbf{G}     & 81.5 & 33.4 & 72.4 & 7.9 & 0.2 & 20.0 & 8.6 & 10.5 & 71.0 & 68.7 & 51.5 & 18.7 & 75.3 & 22.7 & 12.8 & 28.1 & 36.5 &42.7\\
& CRST~\cite{zou2019confidence}       & ICCV 2019   & \textbf{O} & 67.7 & 32.2     & 73.9    & 10.7 & 1.6   & 37.4 & 22.2          & 31.2         & 80.8       & 80.5 & 60.8   & 29.1  & 82.8 & 25.0   & 19.4     & 45.3    & 43.8 &50.1 \\
\midrule
PSPNet                      & DCAN~\cite{wu2018dcan}              & ECCV 2018   & \textbf{G}      & 82.8 & 36.4 & 75.7 & 5.1 & 0.1 & 25.8 & 8.0 & 18.7 & 74.7 & 76.9 & 51.1 & 15.9 & 77.7 & 24.8 & 4.1 & 37.3 & 38.4 &44.9\\
\bottomrule
\end{tabular}%
}
\label{tab:SYCS}
\end{table*}

\subsubsection{Object Detection}
We compare the detection performance of different methods on Cityscapes$\rightarrow$KITTI and Cityscapes$\rightarrow$Foggy Cityscapes. Each dataset contains bounding boxes of different categories. We use  mean Average Precision~(mAP) as the evaluation metric.

Cityscapes and KITTI are both real-world datasets, but collected from different cities. The scene layouts of the images in the two domains are different, which can test the ability to bridge the domain gap caused by both appearance and scene layout differences. Cityscapes and KITTI only have 5 shared categories in the adaptation setting. Foggy Cityscapes is a synthetic  dataset simulating fog on Cityscapes images. Cityscapes and Foggy Cityscapes have 8 classes of objects; since they have the same scene layouts, this adaptation task focuses on testing the appearance adaptation ability of a DA method. 

\subsubsection{Semantic Segmentation}
We compare the segmentation performance of different methods on GTA$\rightarrow$Cityscapes and SYNTHIA$\rightarrow$Cityscapes. Mean intersection-over-union (mIoU) is utilized as the evaluation metric. 

GTA and SYNTHIA are both synthetic datasets, while Cityscapes is a real-world dataset. Both GTA$\rightarrow$Cityscapes and SYNTHIA$\rightarrow$Cityscapes test the performance of simulation-to-real segmentation adaptation methods. Images in GTA and Cityscapes are taken from the dashcams, while images in SYNTHIA are taken from various points of view. Images in GTA have higher level of fidelity compared to images in SYNTHIA. Consequently, SYNTHIA has a larger domain gap than Cityscapes, and it can also test the adaptation method on the domain gap caused by different point of view angles. 

\subsubsection{Result Analysis}

All the experiment result comparisons are shown in Table~\ref{tab:digits}, \ref{tab:office-31}, \ref{tab:office-home}, \ref{tab:visda-2017} (image classification); Table~\ref{tab:CSKITTI},~\ref{tab:CSFCS} (object detection); and Table~\ref{tab:GTACS}, \ref{tab:SYCS} (semantic segmentation). For each backbone, the methods are sorted by average classification accuracy, mAP and mIoU. 

The results show that, compared with object detection and semantic segmentation, it is easier for the methods under analysis to achieve better performance on the image classification task. 
Since classification is a relatively simple task, not requiring many local details for the global class prediction, no specific category  performs significantly better than the others. 
For object detection and semantic segmentation, most of the published work utilize adversarial discriminative or adversarial generative methods since these two tasks require massive detailed local information about the images.
Adversarial learning-based methods are powerful in performing local feature alignment while discrepancy-based and self-supervision-based methods are less capable of capturing local information, leading to inferior performance on object detection and semantic segmentation tasks. 



\section{Future Directions}
\label{sec:NewPerspectives}

Existing DUDA methods have achieved promising performance on many computer vision tasks, such as object classification and semantic segmentation. However, there is still a large performance gap between existing methods and the upper bound~(train and test both on target domain). To help address the remaining challenges, we provide some possible improvements over the state-of-the-art methods. In addition,we present more practical settings of DA, new applications of DA and brave new perspectives in DA.


\subsection{New Methodologies of DA}

\textbf{Incorporating Previous Knowledge.}
As domain shift is usually caused by the imaging process, such as illumination changes~\cite{patel2015visual}, incorporating prior knowledge into the adaptation process may lead to a performance increase. For adversarial methods, imposing multi-level constraints jointly in the adaptation, such as low-level appearances, mid-level features, and high-level semantics, can better preserve the structure and attributes of the source data and thus perform better. 
Designing an effective and direct metric to evaluate the quality of adaptation, instead of testing the performance on the target domain, would accelerate the training process of GANs.

\textbf{Meta Learning Across Domains.}
Meta learning algorithms provides a learning to learn paradigm that is effective at learning meta models with the capability of solving new tasks in a fast manner. However, they require sufficient tasks for meta model training and the optimized model can only solve new tasks similar to the training ones. These limitations make them suffer from performance decline in presence of insufficiency of training tasks in the target domains and task heterogeneity, where the source tasks present different characteristics from the target tasks~\cite{kang2018transferable}. Besides the above challenge, there may be data distribution shift between the source tasks and target tasks, exposing more severe challenges to existing meta learning algorithms. Cross-domain meta learning provides promising solutions to address these challenges by essentially learning more transferable representations~\cite{li2018learning,balaji2018metareg}.

\textbf{Contrastive Learning for DA.}
DUDA methods~\cite{peng2019federated, peng2019domain} are recently focusing on the disentanglement~\cite{lee2018diverse} of the features into domain-invariant and domain-specific ones based on data variations. Domain-invariant features play an important part in reducing the noisy information from each domain, thus making learned features discriminative of the category. 
Current approaches of contrastive learning for domain adaptation are highly dependent on the design of specific tasks. For example, different DA tasks may rely on different pretext tasks. Therefore, a potential research direction is to design a common pretext task. 
These methods are often criticized for their computational cost since a large number of negative samples have to be selected for comparison with every single positive sample. Thus, an approach to decrease computational complexity is needed. 

\subsection{More Practical Settings of DA}

\textbf{Multi-modal DA.}
The labeled source domains may contain multi-modal data. For example, synthetic data generated by simulators (CARLA and GTA-V, etc.) may be of different modalities, such as LiDAR, RaDAR, and image. Other examples include the audio channel and visual channel of videos and the textual and visual information of social posts. Similar to multi-modal recognition~\cite{soleymani2012multimodal,shekhar2013joint} and feature-level fusion in image classification and retrieval~\cite{gao20123,zhao2014affective}, we believe that jointly combining and fusing different modalities to explore the combinations would boost the performance of DA. Another advantage of multi-modal DA is that even if some modalities are missing, the DA system can still work by leveraging information from other available modalities. For example, while the cameras for autonomous driving cannot capture images well at night, the LiDAR scanners are robust under almost all lighting conditions~\cite{wu2017squeezeseg}. How to design effective fusion strategies is the main challenge. The simplest ways are to directly employ early fusion at the feature level or late fusion at the decision level. But to deal well with the incomplete data issue, fusion at the model-level, such as graph convolutional network~\cite{kipf2017semi}, is probably a better choice.

\textbf{Multi-task DA.}
To the best of our knowledge, all the domain adaptation methods proposed so far only focus on a single task~(\textit{e.g.} semantic segmentation, robotic grasping, image classification) with single-modal input~(\textit{e.g.} images). However, in many scenarios, several tasks need to be performed on the same data simultaneously~(\textit{e.g.} semantic segmentation and traffic sign identification for a driving image). Separately adapting each task would be redundant in terms of computation, since the networks for both models may rely on the same set of features. So how to adapt multiple tasks simultaneously and efficiently is a promising direction to explore. One straightforward solution is to find a common feature representation that is beneficial for all the tasks. In order to guide learning towards an optimal shared feature space, methods based on adversarial learning may be used with novel designs.

\textbf{Continual Learning and Adaptation.}
Many machine learning models (\textit{e.g.} semantic segmentation models) are trained on a fixed dataset and then deployed onto a real system, with the assumption that the data at test time has a similar distribution as the training data. However, this is often not the case. Imagine a segmentation model trained on images taken in the US with mostly sunny weather conditions. The cars with the trained model are sold all over the world, and different cars will be running under different domains~(\textit{e.g.} different cities, weathers, time of day, etc.). In order for the network to perform well all the time, continual learning and adaptation needs to be performed. Basically, the network is expected to have the ability to learn continually from a steam of experiential data, building on what has been learned previously, and adapting to varying new domains~\cite{gama2014survey}. Some methods~\cite{kirkpatrick2017overcoming, zenke2017continual, lee2017overcoming} try to limit the extent of weight sharing across experiences by focusing on preserving past knowledge. A method is proposed in~\cite{wu2019ace} to adapt to changing environments for semantic segmentation. However, the method requires synthesizing new images on the fly, which is not computationally efficient. Methods such as~\cite{wang2018dataset} may be used to find a compact representation of the whole dataset, which may be more efficient to fine-tune the model without forgetting the learned knowledge. Learning representations that are generalizable to different domains could make the network more robust against target domain change. \cite{yue2019domain, kim2020learning} proposed to use style transfer to randomize the input domains for better generalization performance. 

\textbf{Federated Domain Adaptation.}
Data generated by IoT devices poses unique challenges for training machine learning models. Users’ profile data typically contains sensitive information, thus cannot leave its hosting device for the sake of privacy preservation. Due to the growing storage and computational power of these devices and concerns about data privacy, it is increasingly preferable to store them in a decentralized way on individual devices rather than hosting them in a central storage. Federated Learning (FL) provides a privacy-preserving mechanism to leverage such decentralized data and computation resources to train machine learning models. The main idea behind federated learning is to have each node learn on its own local data and not share either the data or the model parameters. FL improves data privacy and efficiency in machine learning performed over networks of distributed devices, such as mobile phones, IoT and wearable devices, etc. While FL achieves better privacy and efficiency, existing methods ignore the fact that the data on each node are collected in a non-i.i.d  manner, leading to domain shift between nodes~\cite{peng2019federated}. Models trained with federated learning can still fail to generalize to new devices due to the problem of domain shift. Thus it is of great importance to develop domain adaptation algorithms for federated learning~\cite{peterson2019private}. Such algorithms should be able to align the representations learned among different source and target devices.

\textbf{DA on the Edge.}
Nowadays, many vision-based perception models are deployed in edge devices, \textit{e.g.} mobile phones, autonomous cars, and security cameras. These edge devices are usually deployed in different environments, with substantial need for domain adaptation. Different networks need to be personalized via learning on the users' private data. Sending all the user data to the server, and training millions of networks for all the users would be very expensive. Instead, training networks on the device not only decreases computational complexity, but it also protects privacy since the collected data need not leave the device.
While the edge devices normally have a limited budget in terms of  computation and power, almost all the current state-of-the-art DA methods, \textit{e.g.} the adversarial generative methods, require training on high-end GPUs for days. 
Invertible neural networks~\cite{jacobsen2018revnet, ardizzone2019analyzing} are beneficial to mitigate the memory limitation problem. Other methods, such as quantization, pruning, neural architecture search, and software-hardware co-design, can also be used for efficient on-device training. Performing DA on the edge with efficient deep learning techniques is a practical and fruitful research area to explore.

\subsection{New Applications of DA}
\textbf{Robotics.} Reinforcement learning (RL) algorithms are typically trained in simulation environments. There are two main reasons for this: first, RL algorithms normally require many interactions with the environment, while getting data from the real-world is relatively slow compared to simulation environments that can be sped up; second, training an agent in the real-world would damage the environment as well as the agent itself especially when the policy is not fully well learned. However, if we want to apply the policy learned in simulation into real-world, the domain difference needs to be handled. Methods such as domain randomization~\cite{tobin2017domain} have been proposed to mitigate the visual difference of the domains. Normally, the source and target environments/domains are similar in terms of dynamics of both the environments and agents. An interesting direction is how to transfer if we know the detailed difference of the dynamics.


\textbf{Video Analysis.} Current methods mainly focus on adapting images from the source domain to the target domain. Adapting videos is more challenging and worth studying. Effectively exploring the temporal correlation of videos may significantly improve the DA performance. Existing video style transfer methods \cite{vondrick2016generating,tulyakov2018mocogan,wang2018vid2vid} may fail to work for DA, since the semantics of generated videos cannot be guaranteed to be preserved. Imposing some semantic constraint may help to solve this problem. Further, maintaining the temporal consistency~\cite{lai2018learning} is an important factor. Audio is also an important channel in videos, which is not considered in these methods. 

\textbf{Subjective Attributes.} Existing DA methods work on objective tasks, such as object classification and semantic segmentation, while the adaptation for the understanding of subjective attributes, such as personality~\cite{vinciarelli2014survey}, aesthetics and emotions~\cite{joshi2011aesthetics,zhao2018affective}, has been rarely explored. There are many other challenges to adapt these subjective attributes. Take visual emotion for example: although the transferred images with pixel-level alignment may not change the high-level semantics, the emotion may still be changed~\cite{zhao2018emotiongan,zhao2019cycleemotiongan}. Employing emotion-specific distance measure, such as Mikels' emotion wheel~\cite{zhao2019cycleemotiongan}, may help to tackle this problem. Further, emotions may be evoked by different features, such as low-level artistic elements (\textit{e.g.} color) for abstract paintings and high-level semantics for natural images~\cite{zhao2014affective}. First determining the image style and then conducting adaptation with corresponding semantic consistency may perform better.

\subsection{Brave New Perspectives}

\textbf{DA in the Wild.}
So far all the domain adaptation works mainly focus on a neat setting, however, domain adaptation problems in the real world can be a pretty complex combination of different ``clean'' settings. For example, in a practical domain adaptation setting, there may be several source domains available: some source domains have no labeled examples, some have few labeled samples, and some have abundant labeled samples. At the same time, the label spaces of the source domains and target domains may not be exactly the same. There may also be multiple target domains, with some target domains that have classes not existing in any source domains. Solving practical DA problems in the real-world remains an under-explored field.

\textbf{Model Robustness of DA.}
Due to the wide success of deep neural networks and their unexpected vulnerability of adversarial examples, there has been much attention placed on evaluating and quantifying the robustness of neural networks~\cite{boopathy2019cnn,weng2018towards}. 
However, all the current DA work only focus on boosting the performance on the target domain, without any consideration on the robustness of the adapted model. Investigating how to perform domain adaptation while maintaining the robustness of the model on the target domain is an interesting direction to explore.

\textbf{Neural Architecture Search for DA.}
Existing domain adaptation models usually manually design a specific neural network architecture based the proposed algorithm. However, there is not much work to automatically learn the optimal network architecture to address the domain shift issue. Neural architecture search (NAS)~\cite{wu2019fbnet} is an emerging direction that automatically looks for the optimal neural network architecture for better performance or higher computational efficiency. With the success of NAS, we suggest the research on automatically learning optimal network architecture that can be adapted to different domains. For instance, when detecting vehicles from traffic videos, the model can automatically and dynamically learn different network architectures for videos from different weather, \textit{e.g.}, sunny, rainy, cloudy, and snowy or different locations, \textit{e.g.}, London, New York, Rome and Tokyo. With different network architectures, the model can learn better generalized representation to different domains.

\textbf{Learning Common Sense for DA.}
Most of the existing domain adaptation models try to learn a generalized representation between the source and target domains. However, they do not discover the knowledge behind the visual tasks. We argue that human beings have better domain generalization capability because they can learn the ``common sense'' behind the tasks and infer prediction in different domains. To imitate the human’s capability in domain generalization, we suggest to investigate learning ``common sense'' for domain adaptation. For instance, when the model learns that a computer screen is usually placed on a desk, it can have better performance when detecting the computer screen and the desk, no matter under what illumination, colorization, and camera views. By learning ``common sense'', models can be better generalized to different domains.

\section{Conclusion}
\label{sec:Conclusion}
This paper provides an overview of recent developments in deep unsupervised domain adaptation (DUDA) with the intent of offering a tool for researchers and practitioners to obtain a perspective on the field. Because of the vast literature on the subject, we decided to focus on homogeneous, single-source, single-target, strongly-supervised, and closed-set settings. We classified these methods into different categories, summarized the representative ones, and compared them, supported by experimental results. We believe that DUDA will continue to be an active and promising research area. We also suggested a number of research directions   with a discussion of their main challenges and of some possible solutions.


\ifCLASSOPTIONcaptionsoff
  \newpage
\fi

\bibliographystyle{IEEEtran}
\scriptsize\bibliography{DASurvey_short}


\begin{IEEEbiography}[{\includegraphics[width=1in,height=1.25in,clip,keepaspectratio]{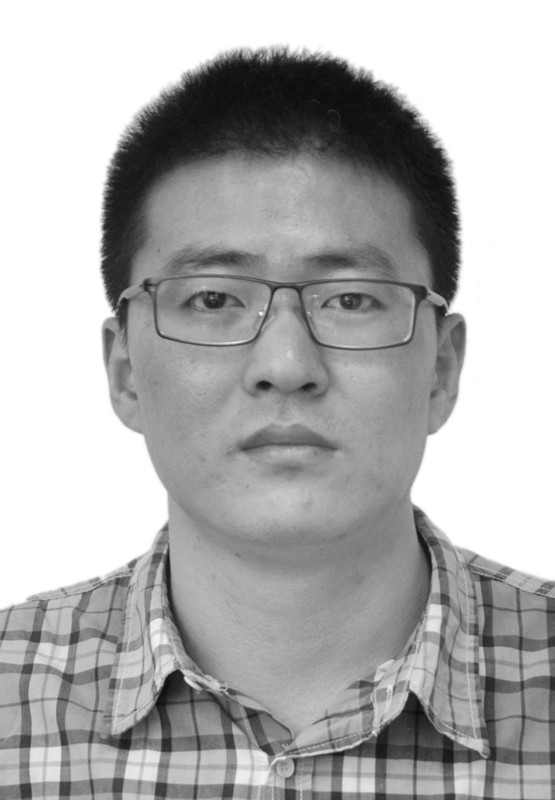}}]{Sicheng Zhao}
(SM'19) received the Ph.D. degree from Harbin Institute of Technology, Harbin, China, in 2016. He has been a Visiting Scholar at National University of Singapore from July 2013 to June 2014, and a Research Fellow at Tsinghua University from September 2016 to September 2017. He is currently a Research Fellow in University of California, Berkeley, USA. His research interests include affective computing, multimedia, and computer vision.
\end{IEEEbiography}

\begin{IEEEbiography}[{\includegraphics[width=1in,height=1.25in,clip,keepaspectratio]{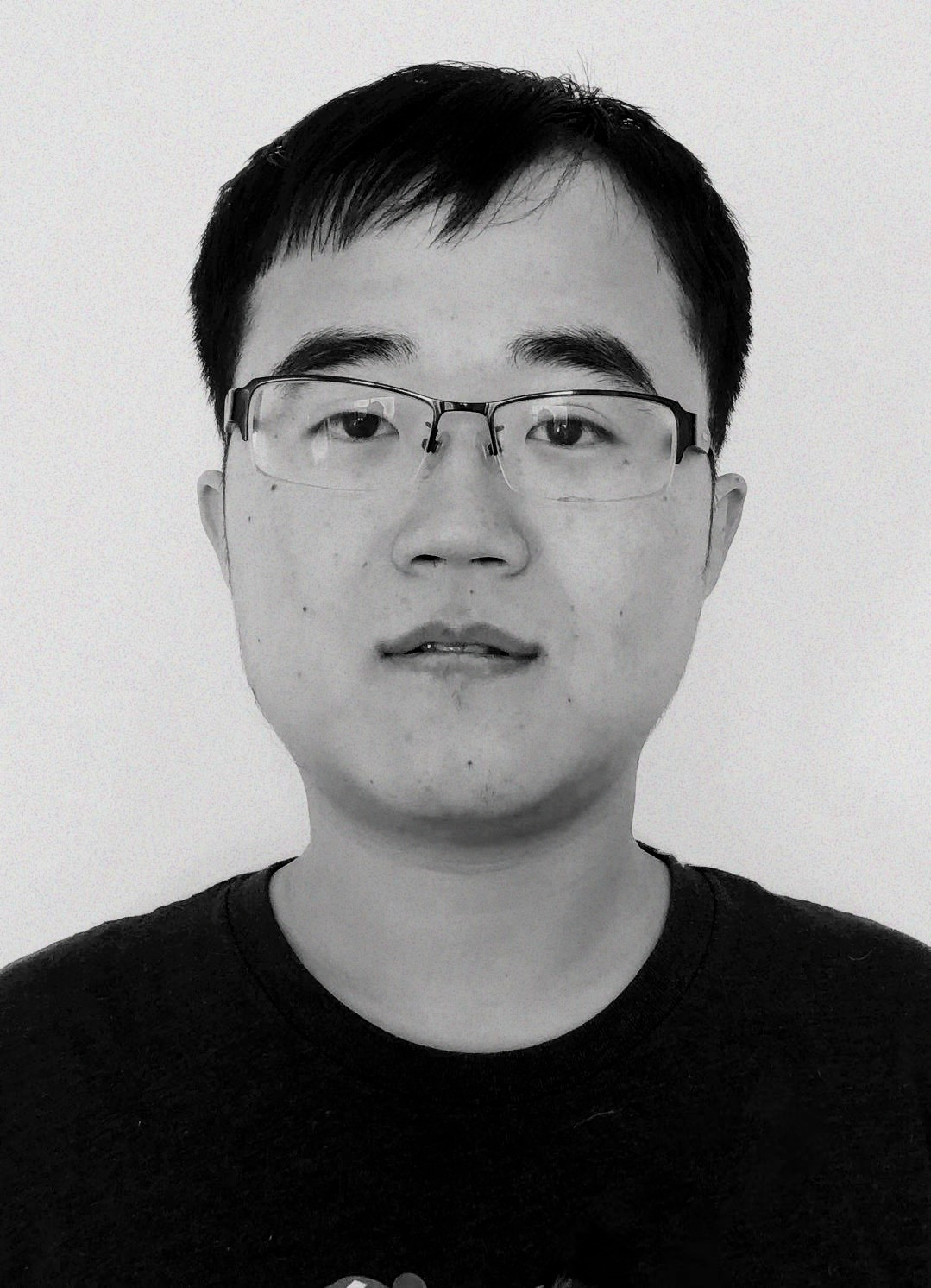}}]{Xiangyu Yue}
received his M.S. degree from Stanford University, Stanford, CA, USA in 2016, and his B.S. degree from Nanjing University, Nanjing, China, in 2014. He is currently a PhD student in University of California, Berkeley, USA. His research interests include computer vision and machine learning.
\end{IEEEbiography}

\begin{IEEEbiography}[{\includegraphics[width=1in,height=1.25in,clip,keepaspectratio]{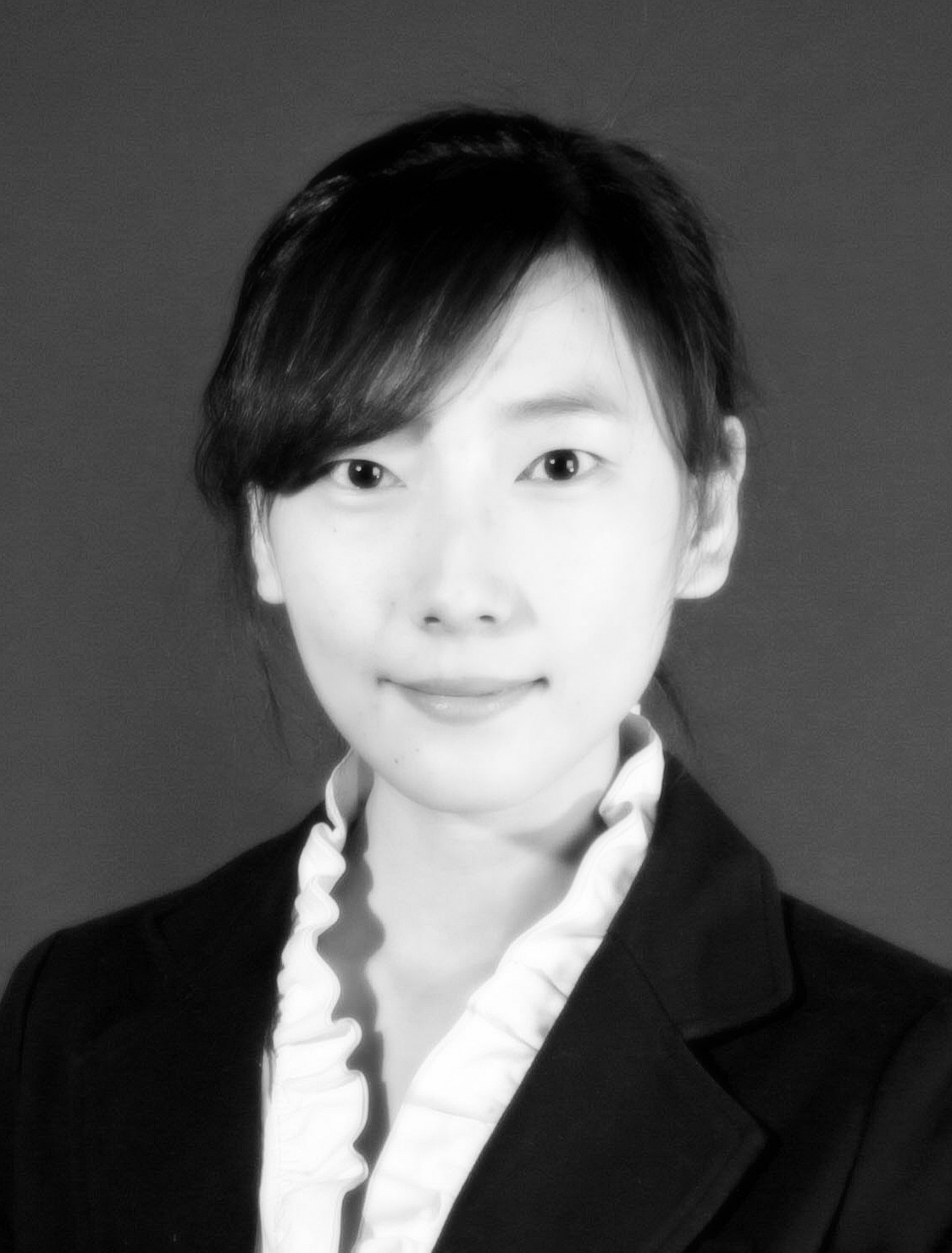}}]{Shanghang Zhang}
received her Ph.D. from Carnegie Mellon University in 2018. She has been selected to “2018 Rising Stars in EECS” USA.  She is currently a Research Fellow in the Berkeley AI Research (BAIR) Lab, the Department of Electrical Engineering and Computer Sciences, University of California, Berkeley, USA. Her research mainly focuses on sample efficient learning, including low-shot learning, domain adaptation, and meta-learning.
\end{IEEEbiography}

\begin{IEEEbiography}[{\includegraphics[width=1in,height=1.25in,clip,keepaspectratio]{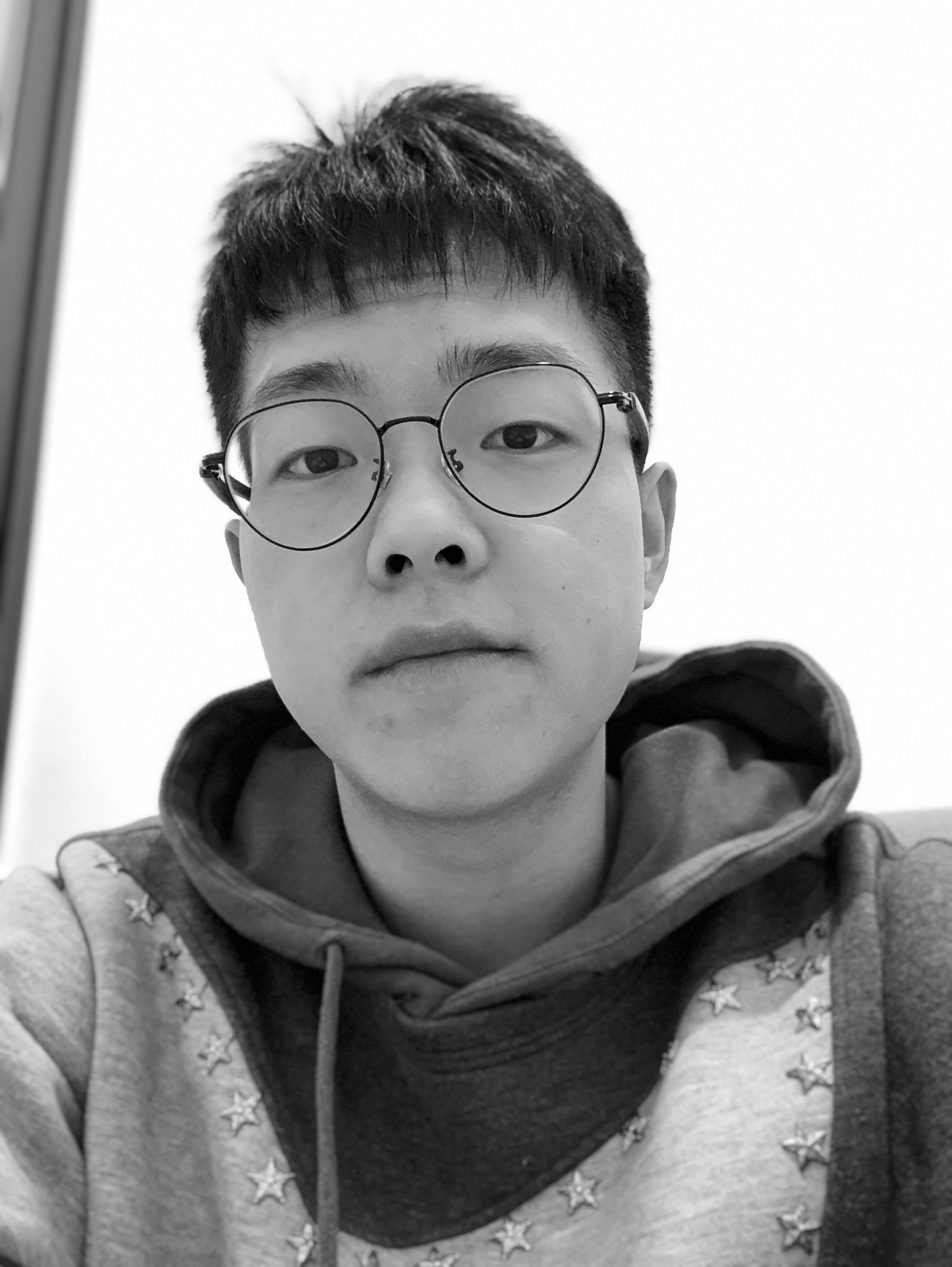}}]{Bo Li}
is currently a research student in University of California, Berkeley, USA and also an undergraduate student at Harbin Institute Technology, China. His research interests include domain adaptation, generative model and variational inference.
\end{IEEEbiography}

\begin{IEEEbiography}[{\includegraphics[width=1in,height=1.25in,clip,keepaspectratio]{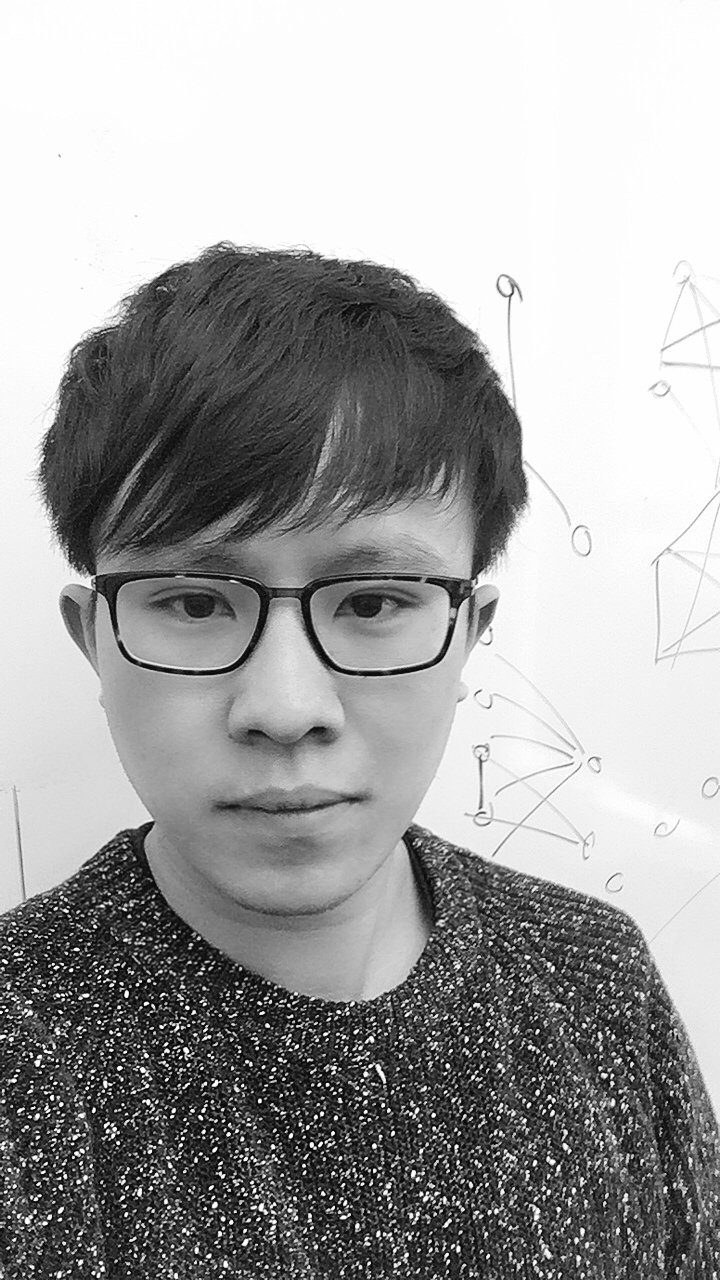}}]{Han Zhao}
received his MMath degree in Mathematics from the University of Waterloo, Canada, in 2015, and his B.S. degree in Computer Science from Tsinghua University, China, 2013. He is currently a Ph.D. candidate at the Carnegie Mellon University, USA. His reseach interests include domain adaptation, probabilistic reasoning and algorithmic fairness.
\end{IEEEbiography}

\begin{IEEEbiography}[{\includegraphics[width=1in,height=1.25in,clip,keepaspectratio]{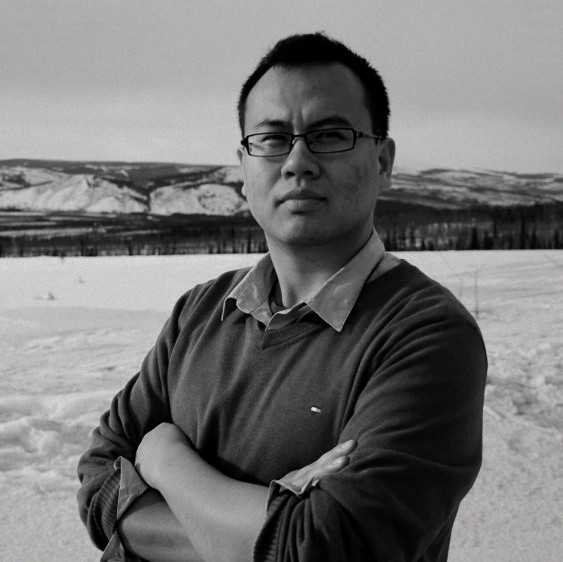}}]{Bichen Wu}
received his his B.S. degree from Tsinghua University, Beijing, China, in 2013. He is currently a PhD student in University of California, Berkeley, USA. His research interests include deep learning, computer vision, and autonomous driving.
\end{IEEEbiography}

\begin{IEEEbiography}[{\includegraphics[width=1in,height=1.25in,clip,keepaspectratio]{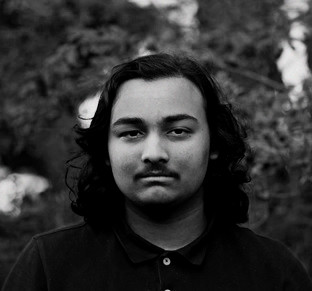}}]{Ravi Krishna}
is currently an undergraduate student at the University of California, Berkeley, USA. His research interests include recommendation systems and natural language processing.
\end{IEEEbiography}

\begin{IEEEbiography}[{\includegraphics[width=1in,height=1.25in,clip,keepaspectratio]{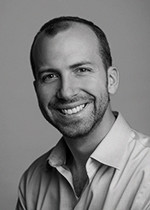}}]{Joseph E. Gonzalez}
received the Ph.D. degree from Carnegie Mellon University, USA, in 2012. He is currently an assistant professor in the EECS department at UC Berkeley and a founding member of the new UC Berkeley RISE Lab. His research interests are at the intersection of machine learning and data systems.
\end{IEEEbiography}

\begin{IEEEbiography}[{\includegraphics[width=1in,height=1.25in,clip,keepaspectratio]{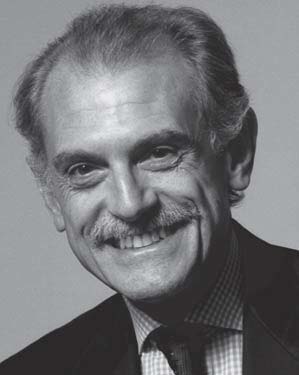}}]{Alberto L. Sangiovanni-Vincentelli}
(F'83) is the Edgar L. and Harold H. Buttner Chair at the Electrical Engineering and Computer
Science Department, University of California at Berkeley, Berkeley, CA, USA, where he has been a member of the faculty since 1976. He helped founding Cadence and Synopsys, the two leading companies in EDA. He
is on the Board of Directors of Cadence, KPIT Technologies, Sonics, Expert Systems, and Cogisen. He is a member of the Investment Committee of Atlante Venture, the advisory board of Walden International and Xseed, and of the Executive Committee of the Italian Institute of Technology. He was the President of the Strategic Committee of the Italian Strategic Fund. He consulted for companies such as Intel, HP, Bell Labs, IBM, Samsung, UTC, Lutron, Camozzi Group, Kawasaki Steel, Fujitsu, Telecom Italia, Pirelli, GM, BMW, Mercedes, Magneti Marelli, ST Microelectronics, ELT, Unipol and UniCredit. He authored over 950 papers, 17 books, and two patents.

Dr. Sangiovanni-Vincentelli is a Fellow of the Association for Computing Machinery (ACM), a member of the National Academy of Engineering, and holds two honorary Doctorates from Aalborg University (Denmark) and KTH (Sweden). He earned the IEEE/RSE Maxwell Award for groundbreaking contributions that have had an exceptional impact on the development of electronics and electrical engineering, the Kaufmann Award for seminal contributions to EDA, the EDAA lifetime Achievement Award, the IEEE/ACM R. Newton Impact Award, the University of California Distinguished Teaching Award, the IEEE Technical Committee on Cyber–Physical Systems (TC-CPS) Technical Achievement Award for pioneering contributions and leadership in cyber–physical systems and design automation, the International Symposium on Physical Design (ISPD) lifetime achievement award, intended for individuals who have made outstanding contributions to the field of physical design automation over multiple decades and the IEEE Graduate Teaching Award for inspirational teaching of graduate students.
\end{IEEEbiography}

\begin{IEEEbiography}[{\includegraphics[width=1in,height=1.25in,clip,keepaspectratio]{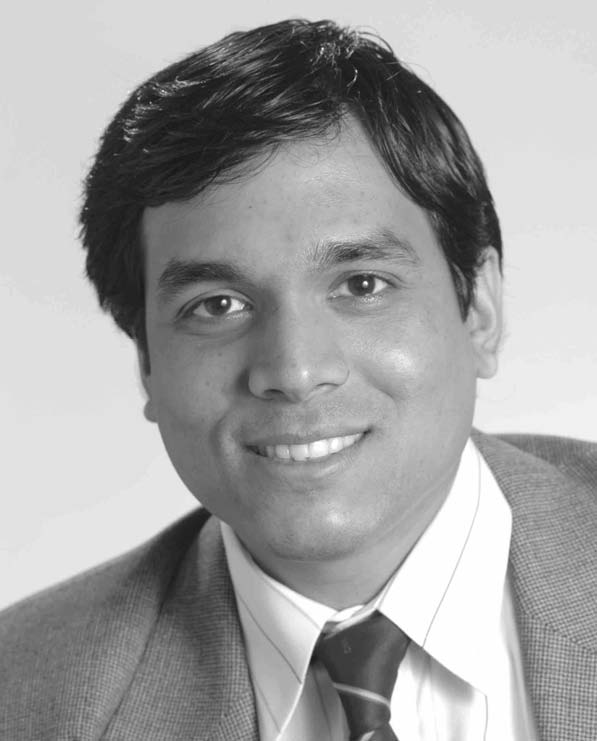}}]{Sanjit A. Seshia}
(F'18) received the B.Tech. degree in computer science and engineering from the Indian Institute of Technology, Bombay, India, in 1998 and the M.S. and Ph.D. degrees in computer science from Carnegie Mellon University, Pittsburgh, PA, USA, in 2000 and 2005, respectively. He is currently a Professor in the Department of Electrical Engineering and Computer Sciences, University of California at Berkeley, Berkeley, CA, USA. His research interests are in dependable computing and computational logic, with a current focus on applying automated formal methods to problems in embedded and cyber–physical systems, electronic design automation, computer security, and
artificial intelligence. His Ph.D. thesis work on the UCLID verifier and decision procedure helped pioneer the area of satisfiability modulo theories (SMT) and SMT-based verification. He is co-author of a widely used textbook on embedded systems. He led the offering of a massive open online course on cyber-physical systems for which his group developed novel virtual lab auto-grading technology
based on formal methods.

Prof. Seshia has served as an Associate Editor of the IEEE TRANSACTIONS ON COMPUTER-AIDED DESIGN OF INTEGRATED CIRCUITS AND SYSTEMS, and as Co-Chair of the Program Committee of the International Conference on Computer-Aided Verification (CAV) in 2012. His awards and honors include a Presidential Early Career Award for Scientists and Engineers (PECASE) from the White House, an Alfred P. Sloan Research Fellowship, the Frederick Emmons Terman Award for engineering education, and the School of Computer Science Distinguished Dissertation Award at Carnegie Mellon University.
\end{IEEEbiography}

\begin{IEEEbiography}[{\includegraphics[width=1in,height=1.25in,clip,keepaspectratio]{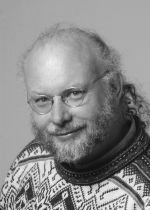}}]{Kurt Keutzer}
(F'96) received his Ph.D.\ degree in Computer Science from Indiana University in 1984 and then joined the research division of AT\&T Bell Laboratories. In 1991 he joined Synopsys, Inc.\ where he ultimately became Chief Technical Officer and Senior Vice-President of Research. In 1998, Kurt became Professor of Electrical Engineering and Computer Science at the University of California at Berkeley. Kurt's research group is currently focused on using parallelism to accelerate the training and deployment of Deep Neural Networks for applications in computer vision, speech recognition, multi-media analysis, and computational finance. Kurt has published six books, over 250 refereed articles, and is among the most highly cited authors in Hardware and Design Automation. Kurt is a Fellow of the IEEE.
\end{IEEEbiography}

\end{document}